\AtBeginDocument{%
  \providecommand\BibTeX{{%
    \normalfont B\kern-0.5em{\scshape i\kern-0.25em b}\kern-0.8em\TeX}}}

\documentclass[sigconf]{acmart}

\makeatletter
\def\@ACM@checkaffil{% Only warnings
    \if@ACM@instpresent\else
    \ClassWarningNoLine{\@classname}{No institution present for an affiliation}%
    \fi
    \if@ACM@citypresent\else
    \ClassWarningNoLine{\@classname}{No city present for an affiliation}%
    \fi
    \if@ACM@countrypresent\else
        \ClassWarningNoLine{\@classname}{No country present for an affiliation}%
    \fi
}
\makeatother

\copyrightyear{2023}
\acmYear{2023}
\setcopyright{acmlicensed}
\acmConference[MM '23] {Proceedings of the 31st ACM International Conference on Multimedia}{October 29--November 3, 2023}{Ottawa, ON, Canada.}
\acmBooktitle{Proceedings of the 31st ACM International Conference on Multimedia (MM '23), October 29--November 3, 2023, Ottawa, ON, Canada}
\acmPrice{15.00}
\acmISBN{979-8-4007-0108-5/23/10}
\acmDOI{10.1145/3581783.3611868}
\usepackage{subfig}
\usepackage{algorithm}
\usepackage{algorithmic}
\begin{document}
\title{FOLT: Fast Multiple Object Tracking from UAV-captured Videos Based on Optical Flow}
\author{Mufeng Yao}
\affiliation{%
  \institution{School of Computer Science,
Shanghai Key Laboratory of Data
Science, Fudan University}
  % \state{Shanghai}
  %\country{China}
}
\email{mfyao21@m.fudan.edu.cn}

\author{Jiaqi Wang}
\affiliation{%
  \institution{School of Computer Science,
Shanghai Key Laboratory of Data
Science, Fudan University}
  % \state{Shanghai}
  %\country{China}
}
\email{21212010033@m.fudan.edu.cn}

\author{Jinlong Peng}
\affiliation{%
  \institution{Tencent Youtu Lab}
  % \state{Shanghai}
  %\country{China}
}
\email{jeromepeng@tencent.com}

\author{Mingmin Chi}
\authornote{Corresponding author}
\affiliation{%
  \institution{School of Computer Science,
Shanghai Key Laboratory of Data
Science, Fudan University}
\institution{Zhengzhou Zhongke Institute of Integrated Circuit and System Application}
  % \state{Shanghai}
  %\country{China}
}
\email{mmchi@fudan.edu.cn}

\author{Chao Liu}
\affiliation{%
  \institution{School of Computer Science,
Shanghai Key Laboratory of Data
Science, Fudan University}
  % \state{Shanghai}
  %\country{China}
  \institution{Zhengzhou Zhongke Institute of Integrated Circuit and System Application}
  % \state{Shanghai}
  %%\country{China}
}
\email{chaoliu@fudan.edu.cn}

\begin{abstract}
Multiple object tracking (MOT) has been successfully investigated in computer vision.
   However, MOT for the videos captured by unmanned aerial vehicles (UAV) is still challenging due to small object size, blurred object appearance, and very large and/or irregular motion in both ground objects and UAV platforms. 
   In this paper, we propose FOLT to mitigate these problems and reach fast and accurate MOT in UAV view.
   Aiming at speed-accuracy trade-off, FOLT adopts a modern detector and light-weight optical flow extractor to extract object detection features and motion features at a minimum cost.
   Given the extracted flow, the flow-guided feature augmentation is designed to augment the object detection feature based on its optical flow, which improves the detection of small objects.
   Then the flow-guided motion prediction is also proposed to predict the object's position in the next frame, which improves the tracking performance of objects with very large displacements between adjacent frames.
   Finally, the tracker matches the detected objects and predicted objects using a spatially matching scheme to generate tracks for every object. 
   Experiments on Visdrone and UAVDT datasets show that our proposed model can successfully track small objects with large and irregular motion and outperform existing state-of-the-art methods in UAV-MOT tasks.
\end{abstract}

\maketitle

\section{Introduction}
\label{chap:introduction}

Multiple object tracking (MOT) aims at identifying objects at each moment from a given video and is widely used in computer vision~\cite{luo2021multiple,2019Deep,2021Analysis}, such as autonomous driving~\cite{geiger2013vision}, human-computer interaction~\cite{chandra2015eye}, and pedestrian tracking~\cite{peng2020chained,2021TrackFormer}.
A common approach in Multiple Object Tracking (MOT) involves two primary stages: detection and association~\cite{shuai2021siammot}. The detection step identifies all objects in each frame, while the association step links objects across consecutive frames to establish complete trajectories for each object~\cite{liu2022multi}.
% A typical MOT method 
% consists of
% two main steps of detection and association~\cite{shuai2021siammot}, with the detection step detecting every object at each frame, and the association step matching objects between adjacent frames to form a complete trajectory for each object~\cite{liu2022multi}.
Recently, MOT in unmanned aerial vehicle (UAV) platforms has attracted extensive research interest.
Compared with conventional MOT~\cite{leal2015motchallenge,dendorfer2020mot20}, MOT in UAV view faces more challenges.

\begin{figure}[htbp]
\setlength{\abovecaptionskip}{0cm}
\setlength{\belowcaptionskip}{0cm}
  \centering
  \includegraphics[width=\linewidth]{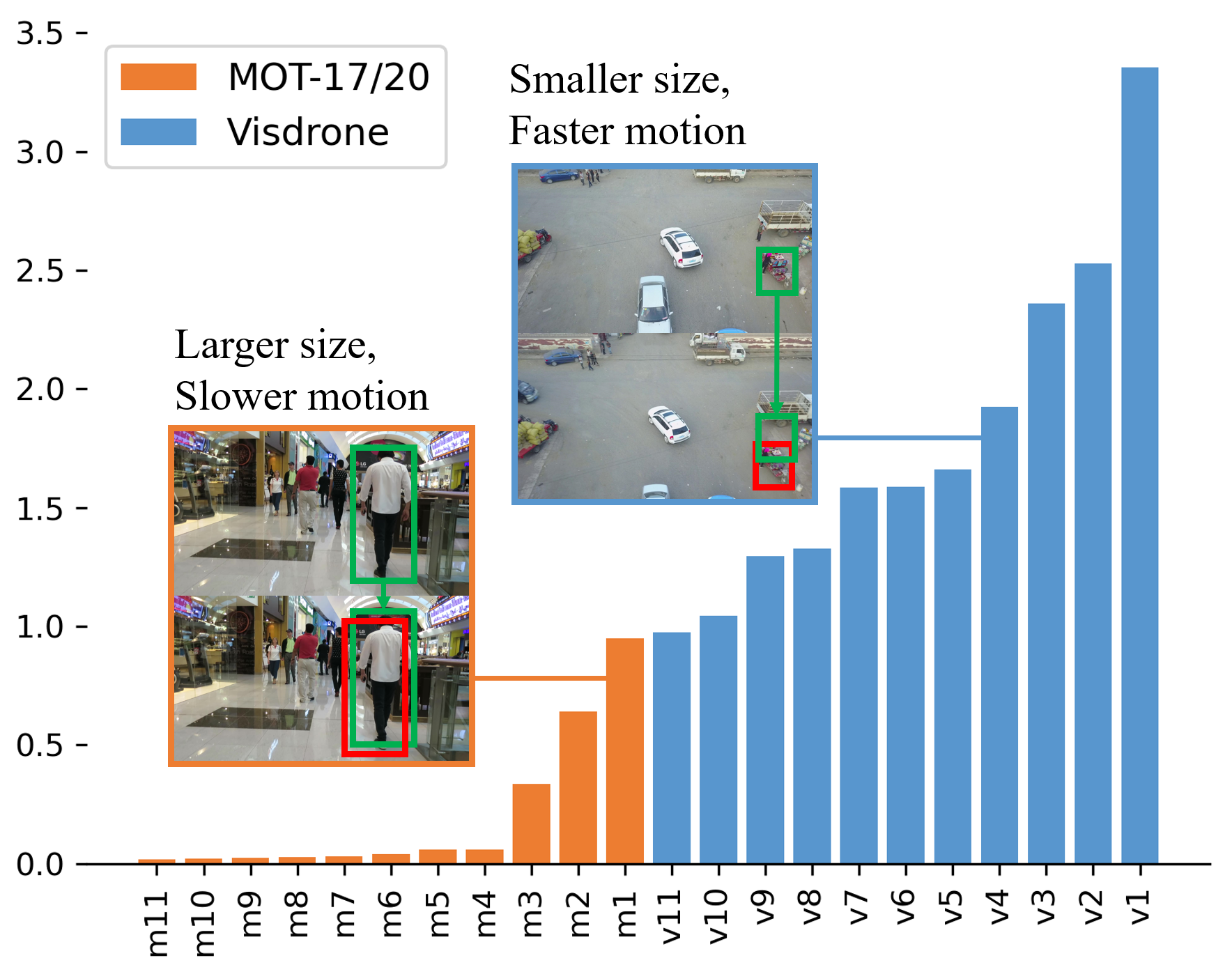}
  \caption{Mean relative accelerations (MRA) of conventional MOT dataset (MOT17/20) and UAV-MOT dataset (Visdrone). All sequences of the MOT17/20 train set (11 in total) and the top 11 sequences of the Visdrone train set are displayed. Green and red rectangle: Visdrone has a smaller object size and faster motion compared with MOT17/20.}
  \label{fig:challenges}
\end{figure}

Firstly, both the ground object and the UAV platform have fast and irregular motion, which makes the tracker difficult to follow the object.
Secondly, the fast and irregular motion will reduce the image quality\cite{kurimo2009effect} and affect the detection of the object.
Thirdly, the object size in aerial view is small, which not only increases the detection difficulty of the detector but also makes the appearance feature of the object unreliable, reducing the accuracy of appearance-matching methods.
We define the mean relative acceleration (MRA) of video sequences in Eq~\eqref{eq:mra} to measure the complexity of object motion.
MRA of a sequence is calculated as the mean of object center acceleration normalized by object size.
As shown in Fig.~\ref{fig:challenges}, the MRA of objects in the Visdrone~\cite{zhu2020detection} dataset is much higher than in the MOT17/20 dataset, showing that object motion patterns in UAV view are more complex than in the conventional MOT task, which requires our tracker modeling the object motion more accurately.
Fig.~\ref{fig:challenges} also indicate that object size in aerial-view videos is smaller compared with traditional street-view videos, which limits the detection performance of the current tracker.

To address these problems, we propose FOLT (Fast Optical fLow Tracker), which utilizes optical flow to model the object motion, augment the object detection feature, and improve tracking performance in UAV view.
The FOLT first use a modern detector and flow-estimator to extract object detection feature and estimate a pixel-wise optical flow map.
Based on the detection features and flow map, the flow-guided feature augmentation is proposed to augment object features at the current frame using the combination of previous features and current optical flow, which improves the detection of small objects.
Then, the flow-guided motion prediction is proposed to model the object's motion and predict its position at a future time, which improves the tracking of objects with large and irregular motion.
Finally, the predicted objects are spatially matched with the detected objects and output tracking results at the current time.

By augmenting the detection feature and explicitly modeling the motion of objects, FOLT can not only improve the detection accuracy but also obtain more stable tracking results in fast-moving scenes of UAV view.
Experiments indicate that our FOLT performs better than previous appearance-matching-based methods in both accuracy and speed, which support our belief that appearance-matching strategy is not important in UAV-MOT scenes.
% In addition, FOLT avoids the problem of unreliable appearance features of small objects since it does not adopt an appearance-matching strategy.

The main contributions of this paper are summarized as follows:
\begin{itemize}
 \item We propose the flow-guided feature augmentation, which combines previous detection features with current features according to optical flow, improves the detection of small objects and mitigates the motion blur problems in UAV-MOT tasks.
     \item We propose the flow-guided motion prediction, which predicts object position according to optical flow, surpassing the commonly used Kalman Filter in both accuracy and speed.
     \item The FOLT proposed based on flow-guided feature augmentation and flow-guided motion prediction reached state-of-the-art on two public MOT-in-UAV-view datasets, which promoted the progress of multiple object tracking in UAV scenes.
\end{itemize}

% \vspace{0.01cm}

\begin{figure*}[t]
\setlength{\abovecaptionskip}{0cm}
\setlength{\belowcaptionskip}{0cm}
\centering
\includegraphics[width=\textwidth]{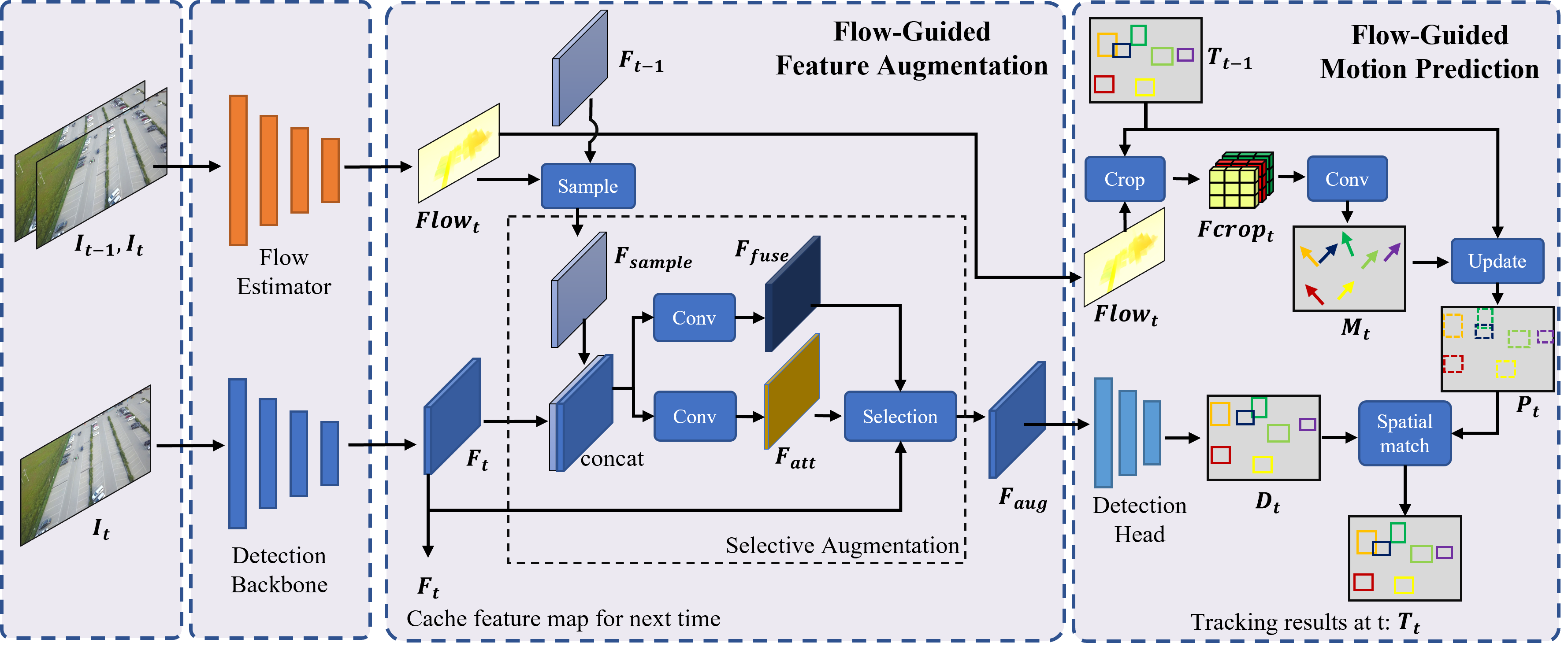}
\caption{The architecture of our proposed FOLT. $I_t, I_{t-1}$ denotes image frame at time $t$ and time $t-1$. $Flow_t, F_t$ denotes the optical flow and detection feature at $t$. $F_{t-1}$ is the cached detection feature of the previous frame. $D_t, M_t, P_t$ denotes the detection results at $t$, the motion prediction of objects at $t$, and the predicted objects position at $t$. $D_t$ and $P_t$ are spatially matched to output $T_t$: the final track results at $t$.}
\label{fig:architecture}
\end{figure*}

\section{Related Work}
\label{chap:related_work}
\subsection{Motion modeling in MOT}
Current MOT methods generally follow the tracking-by-detection scheme~\cite{luo2021multiple,bae2017confidence,fagot2016improving,pirsiavash2011globally,peng2020tpm,peng2020dense},
with the detection step using a deep neural network to output the detection result of each frame,
and the association step completes the inter-frame association based on object appearance~\cite{sugimura2009using,kuo2010multi,li2013survey,yang2012multi,peng2018tracklet} or motion information~\cite{zhao2012tracking,forsyth2006computational,kratz2010tracking,takala2007multi}.
%,yang2012online~\cite{bochinski2017high} 
The association step can be categorized as appearance based~\cite{wojke2017simple,zheng2016mars,wang2020towards,zhang2021fairmot} and motion based~\cite{bewley2016simple,shuai2021siammot,zhou2020tracking,zhang2022bytetrack,bochinski2017high}.
As Fig.~\ref{fig:challenges} and Fig.~\ref{fig:comparisons} show, objects in aerial view are very small and blurry, which will result in unreliable appearance features for the association. Therefore, this paper focuses on motion-based association to avoid unreliable appearance feature problems.
Motion-based association first predicts the position of each object trajectory in the next frame, then it matches the objects detected in the next frame according to their position adjacency, to generate trajectory at the next frame.
The Kalman Filter~\cite{kalman1960new} is the most commonly used motion modeling method and widely adopted by other motion-based trackers~\cite{bewley2016simple,cao2022observation,zhang2022bytetrack}.
SORT~\cite{bewley2016simple} is the classical tracker that applies the Kalman Filter~\cite{kalman1960new} to predict object position.
ByteTrack~\cite{zhang2022bytetrack} makes significant progress by replacing SORT with a more powerful detector and a more complex two-stage matching method.
OCSORT~\cite{cao2022observation} improves the original Kalman Filter in an object-centric manner and performs better in irregular motion scenes.
However, when both objects and the UAV platform are in fast-moving pattern, these Kalman Filter (KF) based methods is hard to follow the large and irregular motion.
In addition to KF based method, SiamMOT~\cite{shuai2021siammot} uses the cross-correlation of the local object regions between adjacent frames to predict the position of the object, CenterTrack~\cite{zhou2020tracking} directly designs a CNN network to predict the object position, UAVMOT~\cite{liu2022multi} uses the topology information of objects in adjacent frames to associate them.
However, these methods introduce too much computation costs and can not reach real-time multiple object tracking.
In this paper, we use a modern light-weight optical flow estimator to realize high speed and accurate motion modeling, which effectively improves tracking in UAV view.
\vspace{0.0cm}
\subsection{Feature fusion in MOT}
Several studies fuse motion information with appearance information in a simple weighted summation scheme\cite{zhang2021fairmot,wang2020towards,wojke2017simple}. 
For example, FairMOT\cite{zhang2021fairmot} compute the cosine distance in feature space and Jaccard distance in image space, then calculate their weighted average value and match the objects according to the averaged distance.
However, these methods didn't fuse information between objects in adjacent frames and are limited in improving object detection on small and blurred objects.
Others use long-short-term memory (LSTM) networks to fuse the temporal features between adjacent frames\cite{milan2017online,xu2020train}.
However, these methods didn't consider motion information when fusing adjacent features.
Different from previous methods, we use optical flow to sample features in their previous position and selectively fuse it with the current feature in an attentive way, which is helpful to objects with small sizes and blurred appearances.
\vspace{0.0cm}
\subsection{Deep flow networks}
Optical flow reflects the relative offset of each pixel between adjacent frames and has a wide range of applications in computer vision tasks~\cite{janai2020computer}, such as autonomous driving~\cite{geiger2012we} action recognition~\cite{simonyan2014two}, and pose tracking~\cite{brox2006high}.
It is feasible to estimate the offset of each object between adjacent frames from a given optical flow map and detected object range.
Traditional optical flow methods~\cite{horn1981determining,sun2014quantitative} have a large computational overhead and are difficult to apply in real time.
Several optical flow neural networks\cite{dosovitskiy2015flownet,hur2019iterative,sun2018pwc} have achieved promising results, providing high accuracy dense optical flow in real-time.
However, these networks are still not fast enough to ensemble in a real-time multiple object tracker, since the detection and association have already cost some computation.
Recently, the FastFlowNet\cite{kong2021fastflownet} reached a comparative flow estimation accuracy while only costing 11ms per image in inference.
Hence, it is possible to integrate this optical flow network into our tracking algorithm and utilize the extracted flow to enhance our object detection feature and predict the object motion.

\section{Proposed Method}
\label{chap:proposed_method}
\subsection{Overview}
Different from conventional MOT tasks, MOT in the UAV platform faces more challenges such as small object size, similar and blurred object appearance, and large and irregular motion.
We define mean relative acceleration (MRA) to reveal this phenomenon:
\begin{equation}
\label{eq:mra}
    MRA = \frac{1}{n}\sum_{i=1}^{n-1}(V(t_{i+1}) - V(t_{i})),
\end{equation}
among them, n is the length of video, $V(t_i)$ is the object's relative velocity at time $i$:
\begin{equation}
    V(t_i) = \frac{\sqrt{(cx(t_i) - cx(t_{i-1}))^2 + (cy(t_i) - cy(t_{i-1}))^2}}{\sqrt{w(t_i)^2 + h(t_i)^2}},
\end{equation}
where $cx(t_i), cy(t_i)$ denote object center position at time $t_i$, $w(t_i)$, $h(t_i)$ denote object width and height at time $t_i$.
% As Fig~\ref{fig:challenges} shows, MOT in UAV datasets have higher MRA and smaller object sizes than conventional datasets.
Capturing from the aerial top-down view, MOT in UAV datasets have higher MRA and smaller object sizes than conventional MOT datasets, which is summarized in Fig~\ref{fig:challenges}.
The higher MRA means that the object's motion is larger and more irregular, which makes the tracker difficult to follow the object.
In addition, the motion blurring and the small object size also increased the detection difficulty in UAV scenes.
Aiming at these difficulties, we propose FOLT to improve the tracking accuracy of small and blurred objects in large and irregular motion scenes.
As Fig.~\ref{fig:architecture} shows, the proposed FOLT consists of three main components.
The feature extraction component consists of a detection backbone and a flow estimator, with the detection backbone extracting the detection feature at current time $F_t$ and the flow estimator estimating a pixel-wise offset map between the current frame and previous frame $Flow_t$.
The flow-guided feature augmentation first samples previous detection features $F_{t-1}$ at the current position to output $F_{sample}$.
Then the $F_{sample}$ and $F_t$ are concatenated and fed into two branches of convolution to output fused feature $F_{fuse}$ and attention weights $F_{att}$.
Then the $F_{fuse}$ and $F_t$ are selectively fused in the guide of attention map $F_{att}$ to output the augmented detection feature $F_{aug}$.
Finally, the augmented detection feature $F_{aug}$ is fed into the detection head to output detection results: $D_t$.
The flow-guided motion prediction takes optical flow map $Flow_t$ and previously tracked object position $T_{t-1}$ as input and uses a convolution layer to predict the object motion $M_t$, then update the object position prediction $P_t$ using $M_t$. Finally, the predicted object position $P_t$ is matched with detected objects $D_t$ to formulate $T_t$ tracking results at time $t$.
We use a two-stage spatial matching strategy~\cite{zhang2022bytetrack} to match the detection results $D_t$ with predicted position $P_t$.
Among the feature extractor, we select the latest YOLOX~\cite{ge2021yolox} as our object detector and tested three pre-trained optical flow networks: FlowNet~\cite{dosovitskiy2015flownet}, PWCNet~\cite{sun2018pwc}, and FastFlowNet~\cite{kong2021fastflownet} as our optical flow extractor.
As Table~\ref{tab:different_flow} shows, the FastFlowNet achieves the best performance on both tracking accuracy and inference speed. Therefore, we adopt FastFlowNet as our base optical flow extractor in subsequent experiments.
Alg~\ref{alg:FOLT} shows the tracking process of FOLT.
\begin{algorithm}[htb]
    \caption{Tracking process of FOLT}
    \label{alg:FOLT}
    \begin{algorithmic}[1] %[1] enables line numbers
    \REQUIRE ~~ An video sequences $\left\{I_t \in \mathbb{R}^{H\times W\times 3}\right\}_{t=1}^{T}$, detection backbone: $D_b$, flow estimator: $FNet$, detection head: $D_h$, flow-guided feature augmentation: $FGFA$, flow-guided motion prediction: $FGMP$ \\
    \ENSURE ~~ Tracking results $Trk$ \\
        \STATE Let $t=0$,$Trk=\{\}$
        \WHILE{$t<T$}
        \IF {$t==0$}
        \STATE Input: $I_t$
        \STATE Extract detection feature $(F_t^1,F_t^2,F_t^3)=D_b(I_t)$
        \STATE Conduct object detection $D_t=D_h(F_t^1,F_t^2,F_t^3)$
        \ELSE
        \STATE Input: two adjacent frames $I_{t-1},I_t$
        \STATE Extract detection feature $(F_t^1,F_t^2,F_t^3)=D_b(I_t)$
        \STATE Extract flow: $Flow_t=FNet(I_{t-1},I_t)$
        \STATE Augment detection feature with optical flow:$F_{aug}^j=
                FGFA(Flow_t,F_{t-1}^j,F_t^j)$,$j\in\{1,2,3\}$
        \STATE Conduct object detection with augmented features $D_t=D_h(F_{aug}^1,F_{aug}^2,F_{aug}^3)$
        \STATE Predict object motion with FGMP:$M_t = FGMP(Flow_t,Trk_{t-1})$
        \STATE Update tracks: $Trk_t = Trk_{t-1} + M_t$
        \ENDIF
        \STATE Match tracks with current detection: $Trk_t = Trk_t \cup D_t$
        \ENDWHILE
        \STATE \textbf{return} $Trk$
    \end{algorithmic}
\end{algorithm}
%%%%%%%%%%%%%%%%%%%%%%%%%%%%%%%%%%%%%%%
\subsection{Feature extraction}
We use the YOLOX-S\cite{ge2021yolox} backbone to extract detection features for every frame, which adopts a feature pyramid structure to output a three-stage feature map with different spatial resolutions: $F_t^1, F_t^2, F_t^3$.
The spatial resolution of the three-stage feature is downsampled by 8, 16, and 32 compared with the original input image. 
The original flow $Flow_t$ is extracted using FastFlowNet with a spatial resolution ($512\times384$), which is lower than the input resolution of the detection branch ($1088\times608$) for faster inference speed.
We rescale the flow value when downsampling it, to keep the right spatial relationship between adjacent feature maps:
\begin{equation}
    Flow_t^i(:,:,x) = sxFlow_t^{d}(:,:,x),
\end{equation}
\begin{equation}
    Flow_t^i(:,:,y) = syFlow_t^{d}(:,:,y),
\end{equation}
where $sx, sy$ is the down-sample scale in the x-axis and y-axis, $Flow_t^{d}(:,:,:)$ is the down-sampled flow map that needs to be rescaled.
We only plot one stage of feature augmentation in Fig~\ref{fig:architecture} and Fig~\ref{fig:fgfa} for simplicity.
The flow-guided motion prediction is conducted in a single stage since the previously tracked objects are already merged in one stage.
Therefore we use the original optical flow with resolution $512\times384$ as the input of the flow-guided motion prediction.
\subsection{Flow-guided feature augmentation}
 \begin{figure}[t]
 \setlength{\abovecaptionskip}{0cm}
\setlength{\belowcaptionskip}{0cm}
\centering
\includegraphics[width=0.5\textwidth]{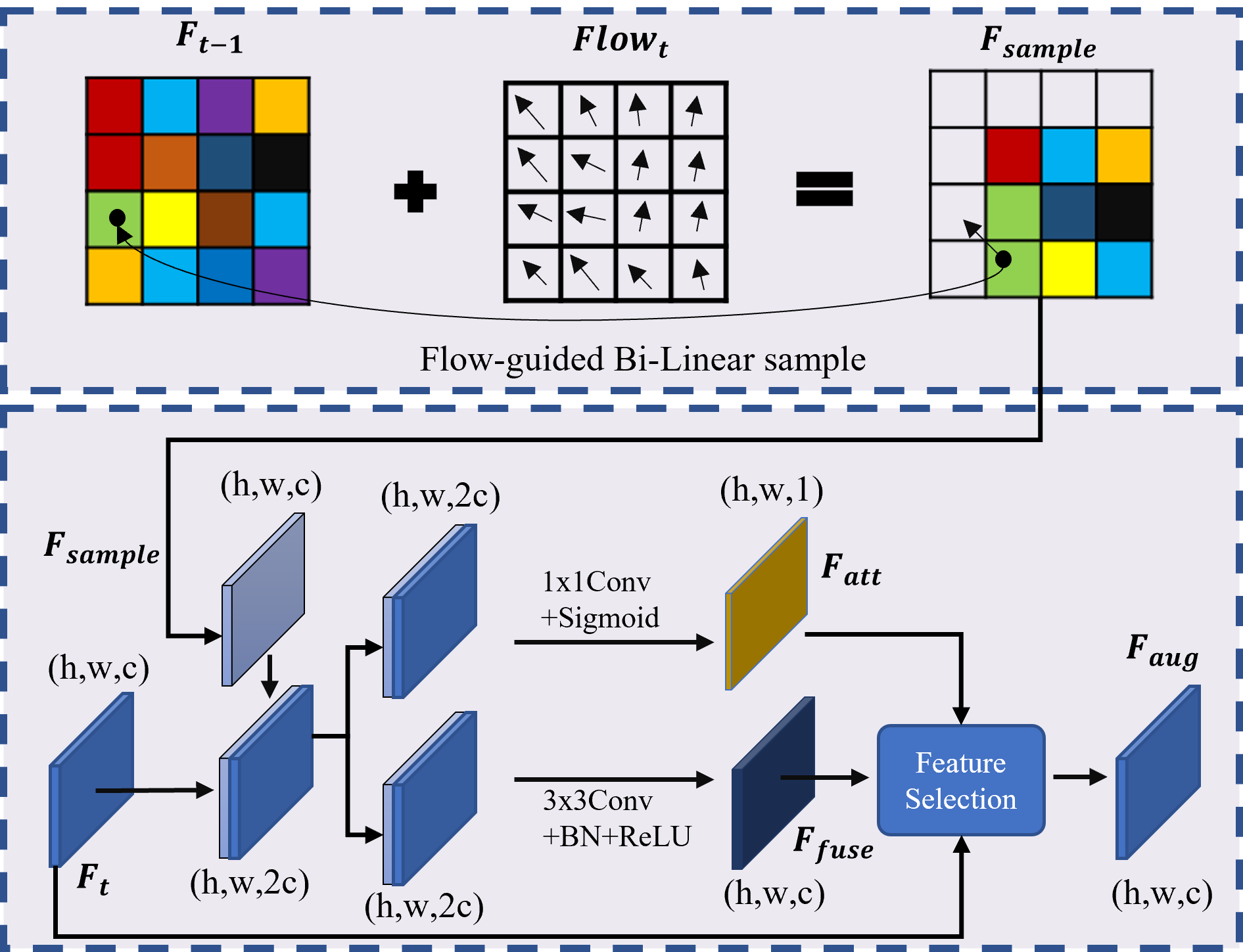}
\caption{Flow-guided feature augmentation. $F_{sample}$ denotes sampled features from the previous frame, $F_t$ denotes current features, $F_{fuse}, F_{att}$ are the fused feature and attention map, $F_{aug}$ is the final augmented features.}
\label{fig:fgfa}
\end{figure}
\vspace{0.01cm}

As Fig~\ref{fig:ablation_fgmp} shows, objects in UAV-captured videos are typically small in size and their appearance is easily affected by the motion blurring, which increases the difficulty of object detection in UAV-view.
Aiming at these challenges, we propose flow-guided feature augmentation (FGFA) to augment object features using previous object features and optical flow between the previous frame and the current frame.
As Fig~\ref{fig:fgfa} shows, our FGFA first use optical flow $Flow_t$ to sample previous detection feature $F_{t-1}$ using a bi-linear sample to output $F_{sample}$:
\begin{equation}
    F_{sample}(i,j) = BS(F_{t-1},i,j,dx,dy),
\end{equation}
where the $dx$ and $dy$ are the optical flow values at position $(i,j)$: $dx = Flow_t(i,j)(x)$, $dy = Flow_t(i,j)(y)$.
The bilinear sample function $BS$ is defined by:
\begin{equation}
\begin{split}
    BS(F,i,j,dx,dy) = s_xs_yF(x,y) + s_y(1-s_x)F(x+1,y) \\
    + (1-s_y)s_xF(x,y+1) + (1-s_y)(1-s_x)F(x+1,y+1),
\end{split}
\end{equation}
where $x=\left[i+dx\right]$, $y=\left[j+dy\right]$, $\left[\cdot\right]$ denote round down operation, $s_x=i+dx - \left[i+dx\right]$, $s_y=j+dy - \left[j+dy\right]$, $F$ is the feature map.
After the flow-guided feature sampling, the $F_{sample}$ and $F_t$ are concatenated in channel dimension and fed into two convolution branches: the attention branch and the fusion branch.
The attention branch takes a $1\times1$ convolution to express $2c$ channels to $1$ channel and then uses a sigmoid function to generate attention scores $F_{att}$ for every spatial position.
The fusion branch takes a $3\times3$ convolution following a batch normalization and a ReLU function to generate fused features $F_{fuse}$.
Finally, the fused feature $F_{fuse}$ and original feature $F_t$ are selectively fused according to attention map $F_{att}$ and output the augmented feature $F_{aug}$:
\begin{equation}
    F_{aug} = F_{fuse}*F_{att} + F_t,
\end{equation}
In accordance with the feature pyramid backbone of YOLOX-S, our flow-guided feature augmentation is conducted in three stages separately:
\begin{equation}
    F_{aug}^i = FGFA(F_{t-1}^i,F_t^i,Flow_t^i),
\end{equation}
We obtain $Flow_t^i$ by down-sample the original flow $Flow_t$ to the same spatial resolution with $F_t^i$ using bilinear sampling.
%%%%%%%%%%%%%%%%%%%%%%%%%%%%%%%%%%
\subsection{Flow-guided Motion Prediction}
%%%%%%%%%%%%%%%%%%%%%%%%%%%%%%%%%%%%%%%
\begin{figure}[t]
\setlength{\abovecaptionskip}{0cm}
\setlength{\belowcaptionskip}{0cm}
    \centering
    \includegraphics[width=0.5\textwidth]{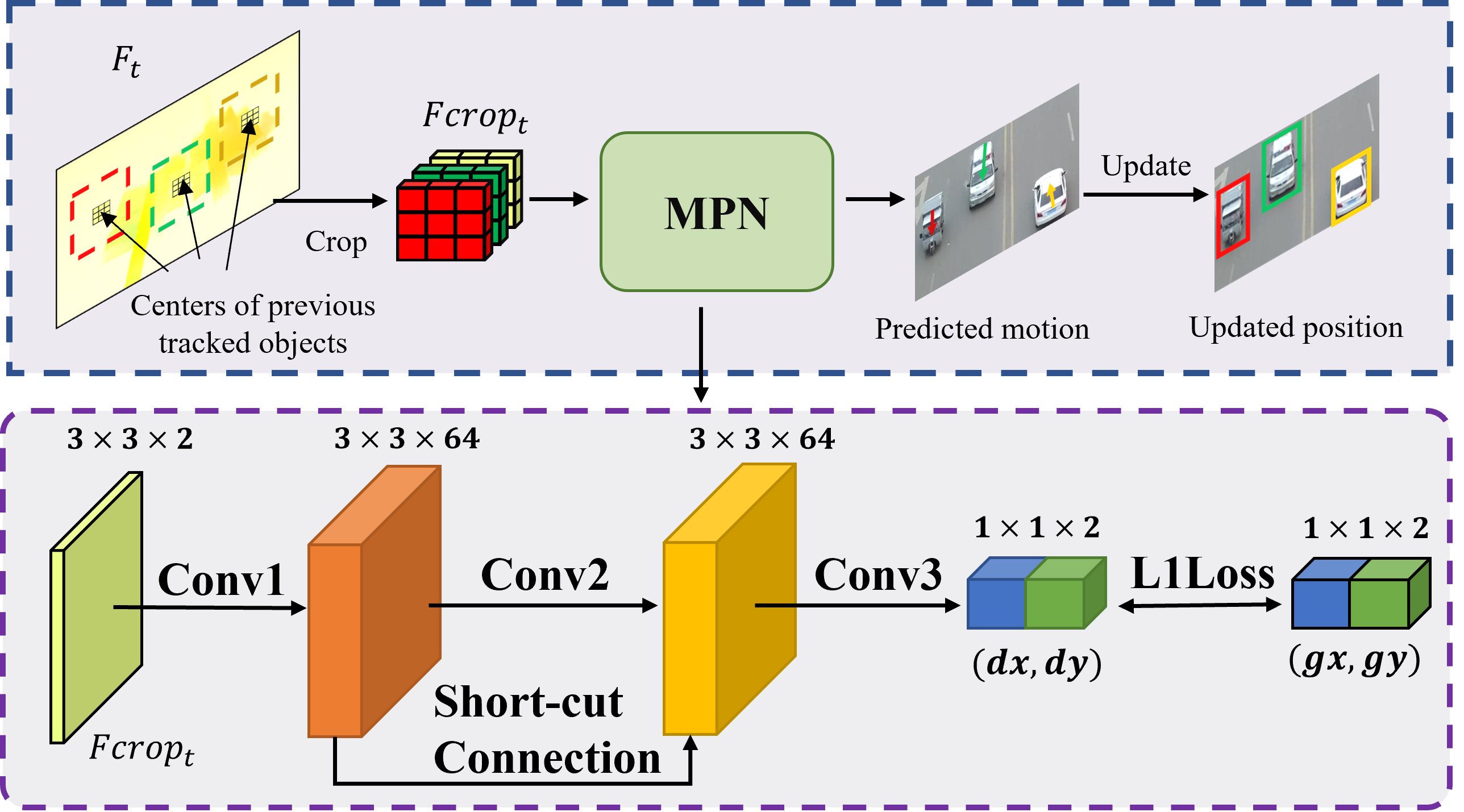}
    \caption{Flow-guided motion prediction. MPN denote motion prediction network, $I_t, I_{t-1}$ denote video frames at time $t,t-1$, $F_t$ denote dense optical flow at time $t$, $Fcrop_t$ denote $3\times3$ pixel of optical flow around the object center, $dx,dy$ denote predicted object center offset at $x,y$ axis, $gx,gy$ denote ground-truth object center offset at $x,y$ axis.}
    \label{fig:fgmp}
\end{figure}
Previous tracking algorithms generally use the Kalman Filter to estimate the object's future position.
However, when both the object and the camera platform have large or irregular motions, the Kalman Filter cannot track the object correctly.
We state that the pixel-wise optical flow map estimated by a deep flow network has already given a rough estimate of the position offset for each pixel.
Therefore, it is practicable to use the statistics of optical flow within each bounding box to estimate its offset between adjacent frames.
Starting from this thought, we propose flow-guided motion prediction to predict every object's motion based on object location and optical flow map.
Fig~\ref{fig:fgmp} shows the architecture of our proposed flow-guided motion prediction.
Given the extracted optical flow map $Flow_t$, we crop the optical flow around the $3\times3$ neighbor of the predicted object center $Fcrop_t$.
Then we feed it to the motion prediction network (MPN) to estimate the object motion between adjacent frames.
As Fig.~\ref{fig:fgmp} shows, the MPN contains 3 layers of convolution, with the second layer having a short-cut connection inspired by ResNet~\cite{he2016deep}.
Each convolution operation is followed by a batch normalization layer and a Softshrink function. Here,
Softshrink is defined by:
\begin{equation}
 {\rm Softshrink}(x) =\left\{
            \begin{aligned}
            & x-\lambda ,\quad x > \lambda &\\
            & 0,\quad -\lambda \leq x \geq \lambda &\\
            & x+\lambda ,\quad x < -\lambda &\\
            \end{aligned}
            \right
            .,
\end{equation}
where $\lambda$ is set to 0.5 in the experiments to ignore too lower values.
The MPN outputs the predicted offset of the object center $M_t=(dx,dy)$.
We use the L1 loss function to supervise the motion prediction:
\begin{equation}
    L1(dx,dy,gx,gy) = |dx - gx| + |dy - gy|,
\end{equation}
where $gx,gy$ is the ground-truth offset of the object in $x$ and $y$ axis. 
The predicted object offset is then added to the previously tracked object position to output the final position estimate:
\begin{equation}
    P_t = T_{t-1} + M_t.
\end{equation}
Finally, the predicted object position $P_t$ is matched with detected objects $D_t$ using a two-stage spatial matching strategy~\cite{zhang2022bytetrack} to generate tracking results at time $t$.
% Experiments show that our proposed flow-guided motion prediction achieves
% higher tracking accuracy than position estimation using Kalman Filter and also reduces the inference time cost.
\begin{table}[t]
\setlength{\abovecaptionskip}{0cm}
\setlength{\belowcaptionskip}{-0.2cm}
\begin{center}
\caption{Effectiveness of different optical flow networks. The best result is marked in bold. The $'\uparrow'$ means that the higher result is better. KF denotes the Kalman Filter, FGMP denotes flow-guided motion prediction.}
\label{tab:different_flow}
    \begin{tabular}{cccccccc}
    \toprule
    Motion modeling & Flow estimator & MOTA$\uparrow$ & IDF1$\uparrow$ & FPS$\uparrow$ \\
    \midrule
    KF & - & 39.6 & 50.4 & 27.0\\
    FGMP & FlowNet-s & 39.9 & 51.3 &  23.2 \\
    FGMP & PWCNet & 40.5 & 54.1 & 15.6 \\
    FGMP & FastFlowNet & \textbf{40.9} & \textbf{55.3} & \textbf{32.0}\\
    \bottomrule
    \end{tabular}
    \end{center}
\end{table}
\vspace{0.001cm}

\begin{table}[t]
\setlength{\abovecaptionskip}{0cm}
\setlength{\belowcaptionskip}{-0.2cm}
\begin{center}
\caption{Ablation studies on Visdrone test-dev set. FGMP indicates flow-guided motion prediction. FGFA indicates flow-guided feature augmentation. The $'\uparrow'$ means that the higher result is better. The best result is marked in bold.}
\label{tab:ablation}
\begin{tabular}{cccccc}
\toprule
Baseline & FGMP    & FGFA    & MOTA$\uparrow$ & IDF1$\uparrow$ & FPS$\uparrow$ \\
\midrule
$\surd$  &         &         & 39.6           & 50.4           & 27.0          \\
$\surd$  & $\surd$ &         & 40.9           & 55.3           & \textbf{32.0}          \\
$\surd$  &         & $\surd$ & 40.9           & 52.3           & 25.5          \\
$\surd$  & $\surd$ & $\surd$ & \textbf{42.1}           & \textbf{56.9}           & 29.4         \\
\bottomrule
\end{tabular}
\end{center}
\end{table}
\section{Experiments}
\label{chap:experiments}
\subsection{Datasets and Metrics}
\textbf{Datasets.} To validate the effectiveness of our proposed FOLT, we conduct comparative experiments on the Visdrone~\cite{zhu2020detection} and UAVDT~\cite{du2018unmanned} datasets.
These two datasets are open source multi-class multi-object tracking datasets, and both are collected from the perspective of UAVs. Therefore, it is suitable to study the tracking objects with small sizes and irregular motion problems in these two datasets.

The Visdrone dataset consists of a training set (56 sequences), validation
set (7 sequences), test-dev set (17 sequences), and test-challenge set (16 sequences).
There are 10 categories in the Visdrone dataset: pedestrian, person, car, van, bus, truck, motor, bicycle, awning-tricycle, and tricycle. Each object within the above categories is annotated by a bounding box, category number, and unique identification number.
In experiments of Visdrone, we use the full ten categories in training while only using five categories in testing,
i.e., car, bus, truck, pedestrian, and van in evaluation, as the evaluation toolkit offered by Visdrone officials only evaluating in these five categories.

UAVDT dataset is a car-tracking dataset in aerial view, it includes different common
scenes, such as squares, arterial streets, and toll stations.
UAVDT dataset consists of a training set (30 sequences), and a test set (20 sequences), with three categories: car, truck, and bus. In experiments of UAVDT, all three categories are evaluated using Visdrone's official evaluation toolkits.

\noindent\textbf{Metrics.} 
To evaluate our tracking efficiency in MOT tasks, we select MOTA and IDF1 as our main evaluation metrics.
The MOTA is calculated as:
\begin{equation}
    MOTA = 1 - \frac{FP + FN + IDs}{GT},
    \label{equ:mota}
\end{equation}
where FP denotes numbers of false positives, FN denotes numbers of false negatives, IDs denote numbers of ID switches, and GT denotes numbers of ground-truth objects.
The IDF1 is calculated as:
\begin{equation}
    IDF1 = \frac{2IDTP}{2IDTP+IDFP+IDFN},
    \label{equ:idf1}
\end{equation}
where IDTP, IDFP, and IDFN denote numbers of true positive, false positive, and false negative that consider ID information.
\begin{figure}[t]
\setlength{\abovecaptionskip}{0cm}
\setlength{\belowcaptionskip}{0cm}
\centering
\includegraphics[width=0.45\textwidth]{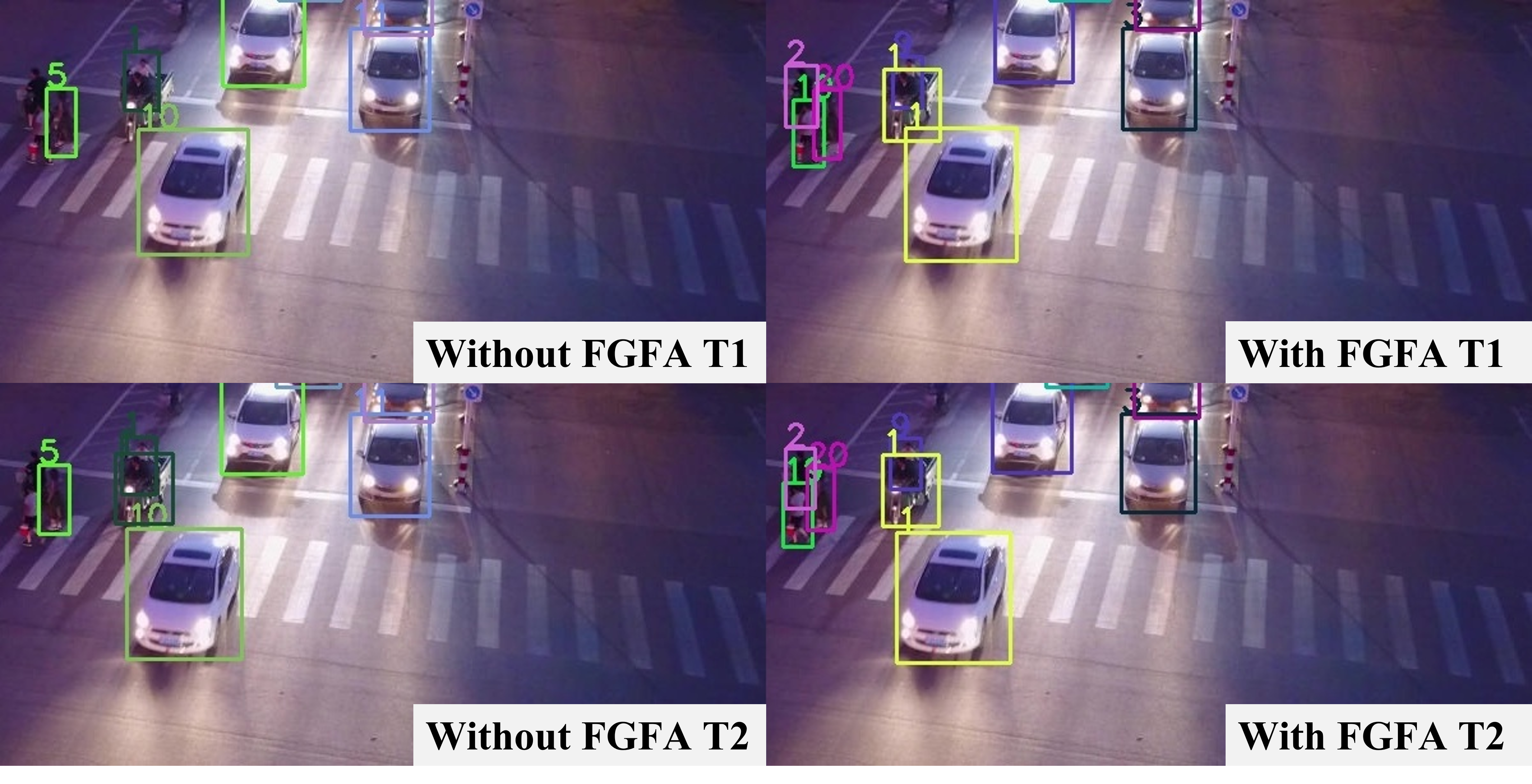}
\caption{Visualization results on small object scenes. The model without FGFA missed the small-sized pedestrains on top-left corner while the model with FGFA tracked them with ID 2,19,20.}
\label{fig:ablation_fgfa}
\end{figure}
\subsection{Implementation Details}
In all experiments, we keep the same train-test split as the official split of the Visdrone and UAVDT datasets.
We use the YOLOX-S\cite{ge2021yolox} model as the base object detector on both datasets, the input image size is $1088\times 608$.
We use the stochastic gradient descent method to optimize the detector, the learning rate is set to $0.000005$, the batch size is set to 4, each dataset is trained for 10 epochs, and the training and testing are completed on a single 2080TI graphic card.
We use L1-Loss to train our motion prediction network and the object localization branch of the detector,
 and use the cross-entropy loss to train the object classification branch of the detector.
In the ablation study, we select MOTA and IDF1 as our evaluation metrics, which are commonly used in MOT tasks.
In comparison with state-of-the-art methods,
we select the accuracy evaluation tool officially provided by the Visdrone dataset to complete the comparison of all metrics.

\subsection{Ablation study}
\begin{table*}[tb]
\setlength{\abovecaptionskip}{0cm}
\setlength{\belowcaptionskip}{-0.2cm}
\begin{center}
    \caption{Comparisons of the proposed FOLT with state-of-the-art methods on Visdrone and UAVDT test sets. The best result is marked in bold. The $'\uparrow'$($'\downarrow'$) means that the higher (lower) result is better.
    Our proposed FOLT surpasses state-of-the-arts in both tracking accuracy and inference speed.}
    \label{tab:sota_comparisons}
    \resizebox{0.9\textwidth}{36mm}{    %40mm
\begin{tabular}{ccc|c|cccrrr}
\toprule
Dataset & Method & Pub $\&$ Year & Speed(FPS)$\uparrow$ & MOTA$\uparrow$ & MOTP$\uparrow$ & IDF1$\uparrow$ & FP$\downarrow$ & FN$\downarrow$ & IDs$\downarrow$ \\
\midrule
         & GOG~\cite{pirsiavash2011globally}   & CVPR2011 & 2.0  & 28.7          & 76.1 & 36.4          & 17706 & 144657 & 1387 \\
         & SORT~\cite{bewley2016simple}        & ICIP2016 & 23.5 & 14.0          & 73.2 & 38.0          & 80845          & 112954 & 3629 \\
         & IOUT~\cite{bochinski2017high}       & AVSS2017 & 27.3 & 28.1          & 74.7 & 38.9          & 36158          & 126549 & 2393 \\
VisDrone & SiamMOT~\cite{shuai2021siammot}     & CVPR2021 & 11.2 & 31.9          & 73.5 & 48.3          & 24123          & 142303 & 862  \\
         & ByteTrack~\cite{zhang2022bytetrack} & ECCV2022 & 27.0 & 35.7          & 76.8 & 37.0          & 21434          & 124042 & 2168 \\
         & UAVMOT~\cite{liu2022multi}          & CVPR2022 & 12.0 & 36.1          & 74.2 & 51.0          & 27983          & 115925 & 2775 \\
         & OCSORT~\cite{cao2022observation}    & CVPR2023 & 26.4 & 39.6          & 73.3 & 50.4          & \textbf{14631}          & 123513 & 986  \\
         & FOLT (Ours)                    & Ours     & \textbf{29.4} & \textbf{42.1}          & \textbf{77.6} & \textbf{56.9}          & 24105          & \textbf{107630} & \textbf{800}  \\
\midrule
         & GOG~\cite{pirsiavash2011globally}   & CVPR2011 & 2.0  & 35.7          & 72   & 0.3           & 62929          & 153336 & 3104 \\
         & SORT~\cite{bewley2016simple}        & ICIP2016 & 23.5 & 39.0          & 74.3 & 43.7          & 33037 & 172628 & 2350 \\
         & IOUT~\cite{bochinski2017high}       & AVSS2017 & 27.3 & 36.6          & 72.1 & 23.7          & 42245          & 163881 & 9938 \\
         & SiamMOT~\cite{shuai2021siammot}     & CVPR2021 & 11.2 & 39.4          & 76.2 & 61.4          & 46903          & 176164 & \textbf{190}  \\
UAVDT    & ByteTrack~\cite{zhang2022bytetrack} & ECCV2022 & 27.0 & 41.6          & 79.2 & 59.1          & \textbf{28819}          & 189197 & 296  \\
         & UAVMOT~\cite{liu2022multi}          & CVPR2022 & 12.0 & 46.4          & 72.7 & 67.3          & 66352          & \textbf{115940} & 456  \\
         & OCSORT~\cite{cao2022observation}    & CVPR2023 & 26.4 & 47.5          & 74.8 & 64.9          & 47681          & 148378 & 288  \\
         & FOLT (Ours)                    & Ours     & \textbf{29.4} & \textbf{48.5} & \textbf{80.1} & \textbf{68.3} & 36429          & 155696 & 338 \\
         \bottomrule
\end{tabular}}
\end{center}
\end{table*}
% We perform our ablation experiments on the Visdrone test-dev set, Table~\ref{tab:ablation} shows the results of the ablation experiments on the Visdrone test-dev set.

\noindent\textbf{Baseline model.} The baseline model we compared with is the model that use YOLOX-S as detector, extended Kalman Filter as motion modeling~\cite{cao2022observation}, and two-stage spatial matching as association~\cite{zhang2022bytetrack}.

\noindent\textbf{Different optical-flow extractor.}
We evaluate the effectiveness of different optical-flow extractors in Table~\ref{tab:different_flow}. As Table~\ref{tab:different_flow} shows, the modern optical flow network FastFlowNet performs better than previous old methods (FlowNet and PWCNet) in both tracking accuracy (40.9 in MOTA and 55.3 in IDF1) and inference speed (32.0 in FPS). Therefore, we adopt FastFlowNet as our base optical flow estimator in subsequent experiments.

\noindent\textbf{Flow-guided feature augmentation.}
In this section, we evaluate the effectiveness of flow-guided feature augmentation (FGFA).
As shown in Table~\ref{tab:ablation_fgfa}, our FGFA performs better than the single frame feature and simple feature warp strategy, with MOTA 1.3 higher than the second best (40.9 to 39.6) and IDF1 1.6 higher than the second best(52.3 to 50.4), while only slower the inference speed a little (27.0 to 25.5 FPS).
Fig~\ref{fig:ablation_fgfa} shows that after using FGFA, our model successfully tracks small pedestrians (with ID 2, 10, and 20), while the model without FGFA fails to track all of them.
Fig~\ref{fig:mota_size} shows that after using FGFA, the tracking improvements (shown in black dash lines) of small objects are larger than normal objects.
In summary, the qualitative and quantitative results above confirm the effectiveness of our proposed FGFA in improving object tracking with small size and motion blur.
\begin{table}[htbp]
\setlength{\abovecaptionskip}{0cm}
\setlength{\belowcaptionskip}{-0.2cm}
\caption{Effectiveness of our flow-guided feature augmentation module. The best result is marked in bold. The $'\uparrow'$ means that the higher result is better.}
\label{tab:ablation_fgfa}
\begin{tabular}{cccc}
\toprule
Motion modeling & MOTA$\uparrow$ & IDF1$\uparrow$ & FPS$\uparrow$  \\
\midrule
Single frame   & 39.6 & 50.4 & \textbf{27}   \\
Naive feature warp    & 39.5 & 50.4 & 26.2 \\
Flow-guided feature augmentation& \textbf{40.9} & \textbf{52.3} & 25.5 \\
\bottomrule
\end{tabular}
\centering
\end{table}
\begin{figure}[t]
\setlength{\abovecaptionskip}{0cm}
\setlength{\belowcaptionskip}{0cm}
\centering
\includegraphics[width=0.4\textwidth]{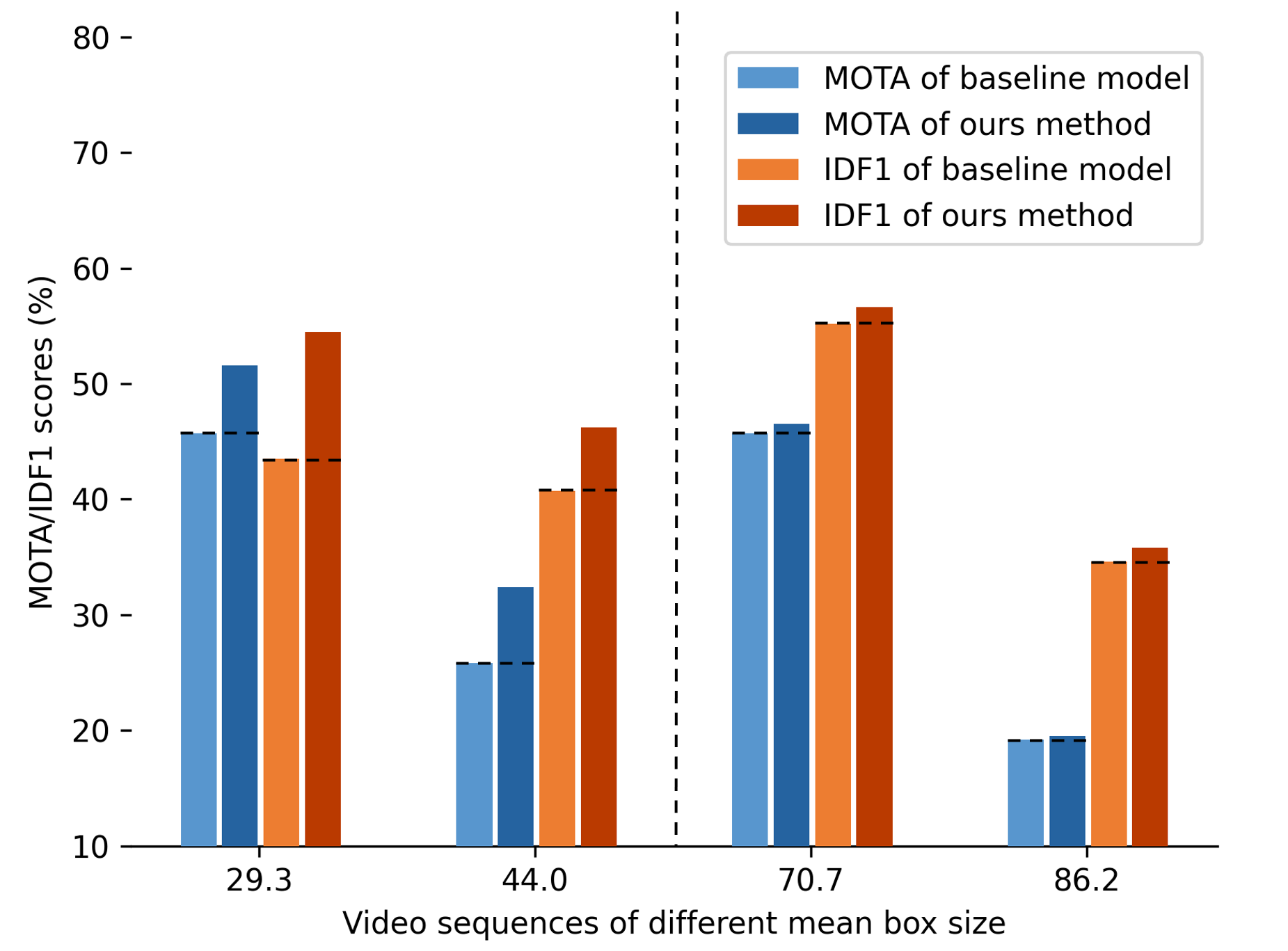}
\caption{Comparisons of baseline and FOLT in Visdrone test-dev set.
The left two groups are the results of two videos with smaller object sizes, while the right two groups are with larger object sizes.
The improvement in MOTA and IDF1 compared with the baseline model are explicitly higher in videos with smaller objects than in videos with larger objects.
 }
\label{fig:mota_size}
\end{figure}

\begin{table}[htbp]
\setlength{\abovecaptionskip}{0cm}
\setlength{\belowcaptionskip}{-0.2cm}
\caption{Effectiveness of our flow-guided motion prediction module. The best result is marked in bold. The $'\uparrow'$ means that the higher result is better.}
\label{tab:ablation_fgmp}
\begin{tabular}{cccc}
\toprule
Motion modeling & MOTA$\uparrow$ & IDF1$\uparrow$ & FPS$\uparrow$  \\
\midrule
Kalman Filter   & 39.6 & 50.4 & 27   \\
Mean of flow    & 40.4 & 53.1 & \textbf{32.3} \\
Flow-guided Motion Prediction& \textbf{40.9} & \textbf{55.3} & 32.0 \\
\bottomrule
\end{tabular}
\centering
\end{table}

 \begin{figure}[t]
\setlength{\abovecaptionskip}{0cm}
\setlength{\belowcaptionskip}{0cm}
\centering
\includegraphics[width=0.5\textwidth]{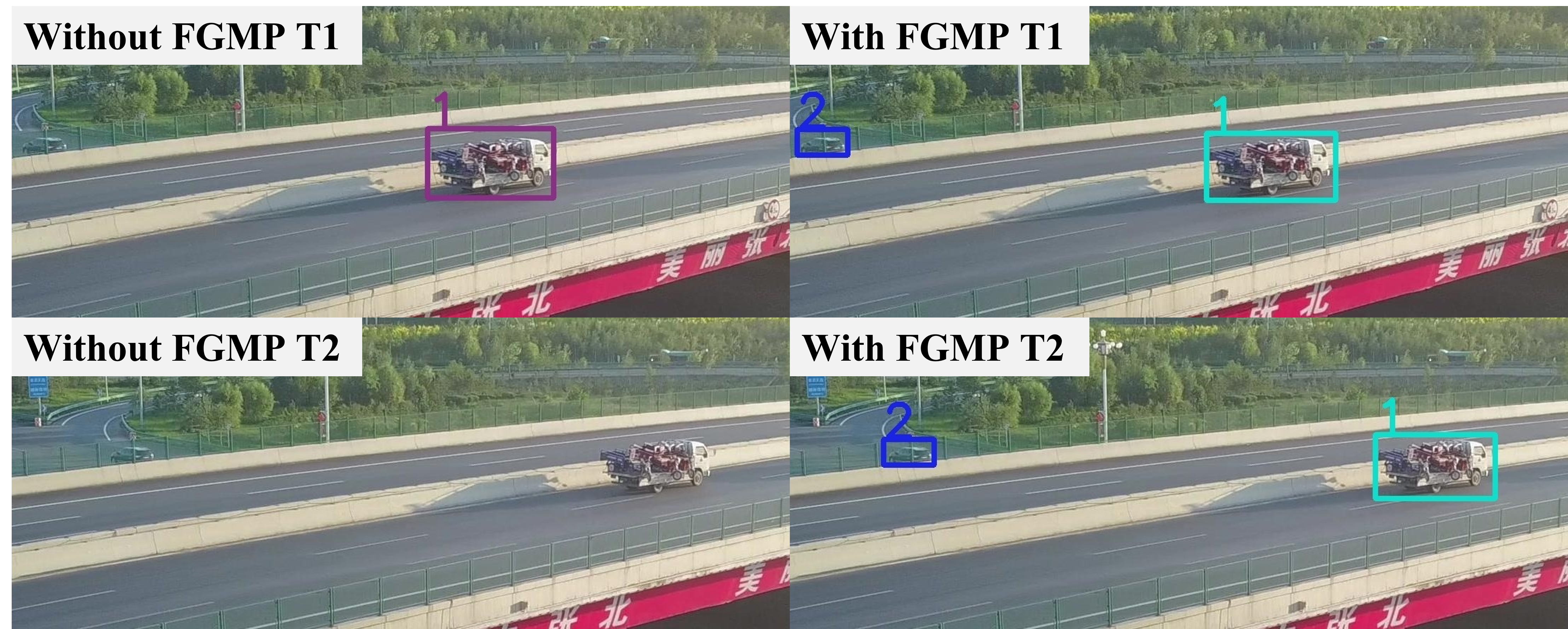}
\caption{Visualization results on large and irregular motion scenes. The model without FGMP missed the fast-moving trucks on the highway at time T while the model with FGMP tracked it with ID 1.}
\label{fig:ablation_fgmp}

\end{figure}
\begin{figure}[t]
\setlength{\abovecaptionskip}{0cm}
\setlength{\belowcaptionskip}{0cm}
\centering
\includegraphics[width=0.4\textwidth]{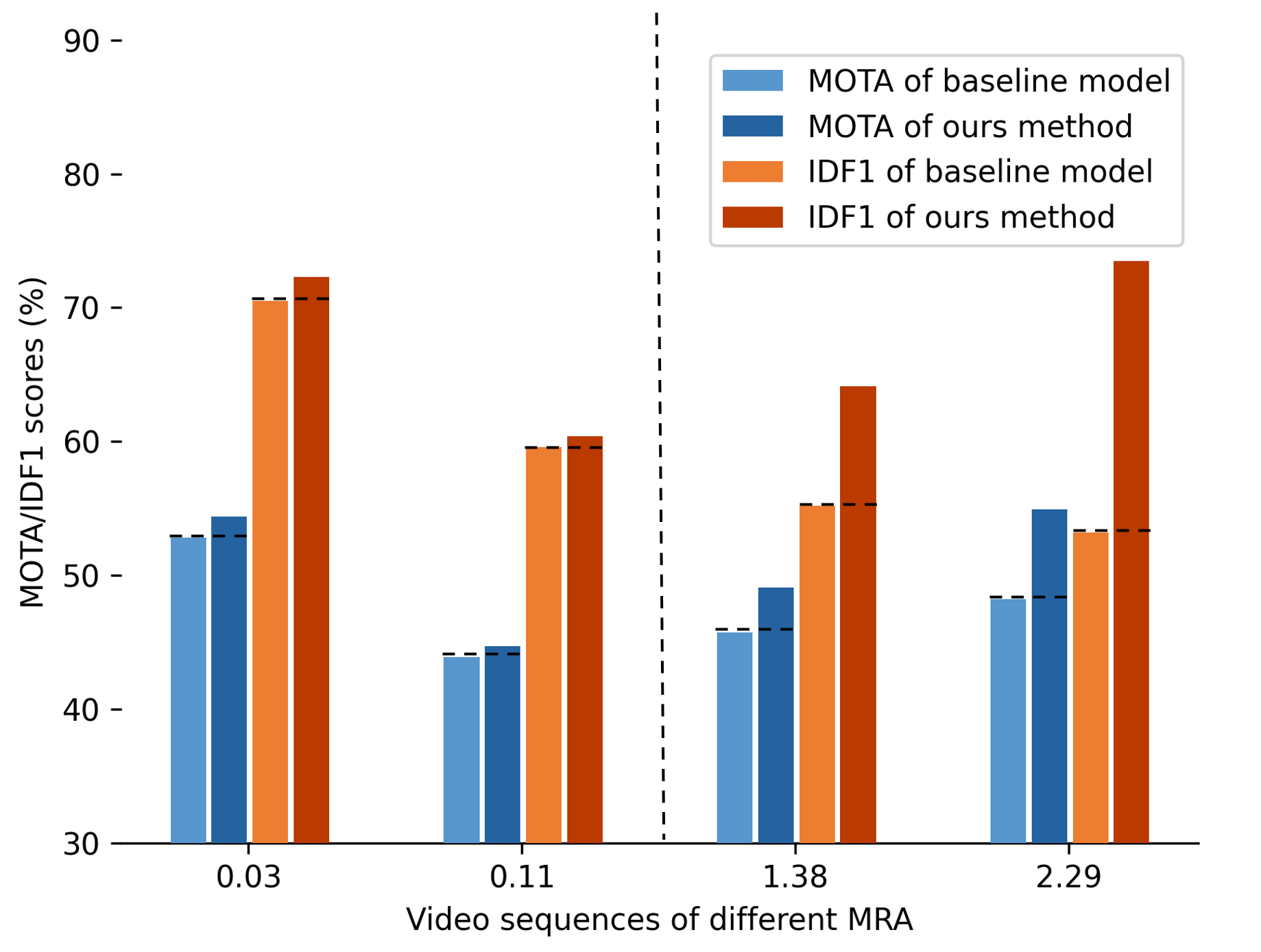}
\caption{Comparisons of baseline and FOLT in Visdrone test-dev set.
The left two groups are the results of two videos with low MRA, while the right two groups are with high MRA.
The improvement in MOTA and IDF1 compared with the baseline model are explicitly higher in videos with high MRA than in videos with low MRA.
 }
\label{fig:mota_bigoffset}
\end{figure}

\noindent\textbf{Flow-guided motion prediction.}
As Table~\ref{tab:ablation_fgmp} shows, compared with the Kalman Filter, using optical flow as a motion estimator not only improves tracking accuracy (39.6 to 40.4 in MOTA and 50.4 to 53.1 in IDF1) but also increases inference speed (27.0 to 32.3 in FPS).
Second, using flow-guided motion prediction (FGMP) further improves tracking accuracy (40.4 to 40.9 in MOTA and 53.1 to 55.3 in IDF1) but only introduces a little speed cost (32.3 to 32.0 in FPS).

As Fig~\ref{fig:mota_bigoffset} shows, after using FGMP in tracking, the improvements on objects with large and irregular motion (higher MRA) are larger than objects with slower motion (lower MRA).
As Fig~\ref{fig:ablation_fgmp} shows, after using our FGMP, the tracker successfully tracks the fast-moving truck on the highway (see the object with ID1), while the model without FGMP fails to track the fast-moving truck at time $t$.
In summary, the qualitative and quantitative results above confirm the effectiveness of our proposed FGMP in improving object tracking with large and irregular motion.
\begin{figure*}[htbp]
\setlength{\abovecaptionskip}{0cm}
\setlength{\belowcaptionskip}{0cm}
\centering
	\subfloat[Visdrone dataset]{\includegraphics[width = 0.48\textwidth]{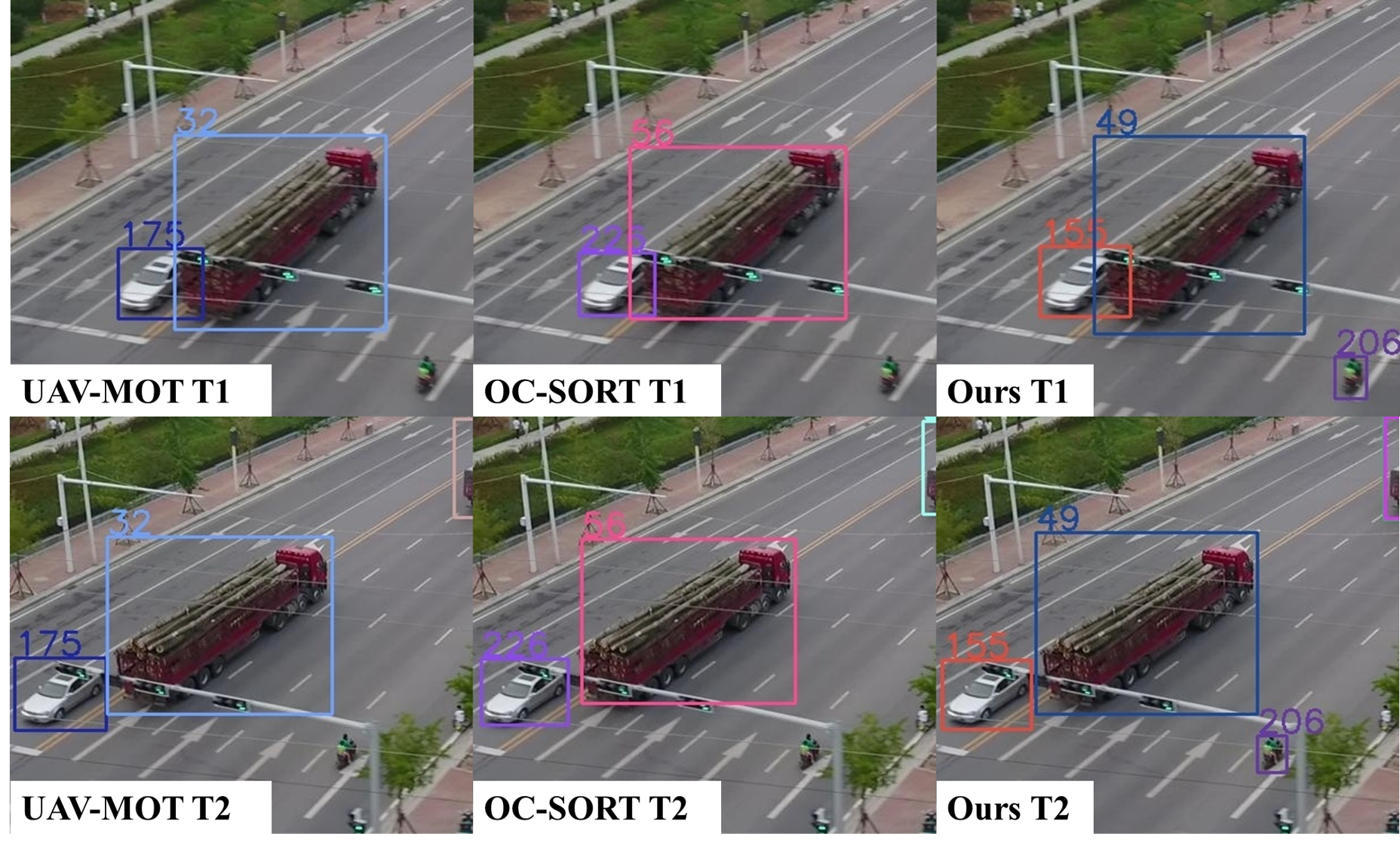}
\label{fig:comparisons:visdrone}}    
	\hfill
	\subfloat[UAVDT dataset]{\includegraphics[width = 0.48\textwidth]{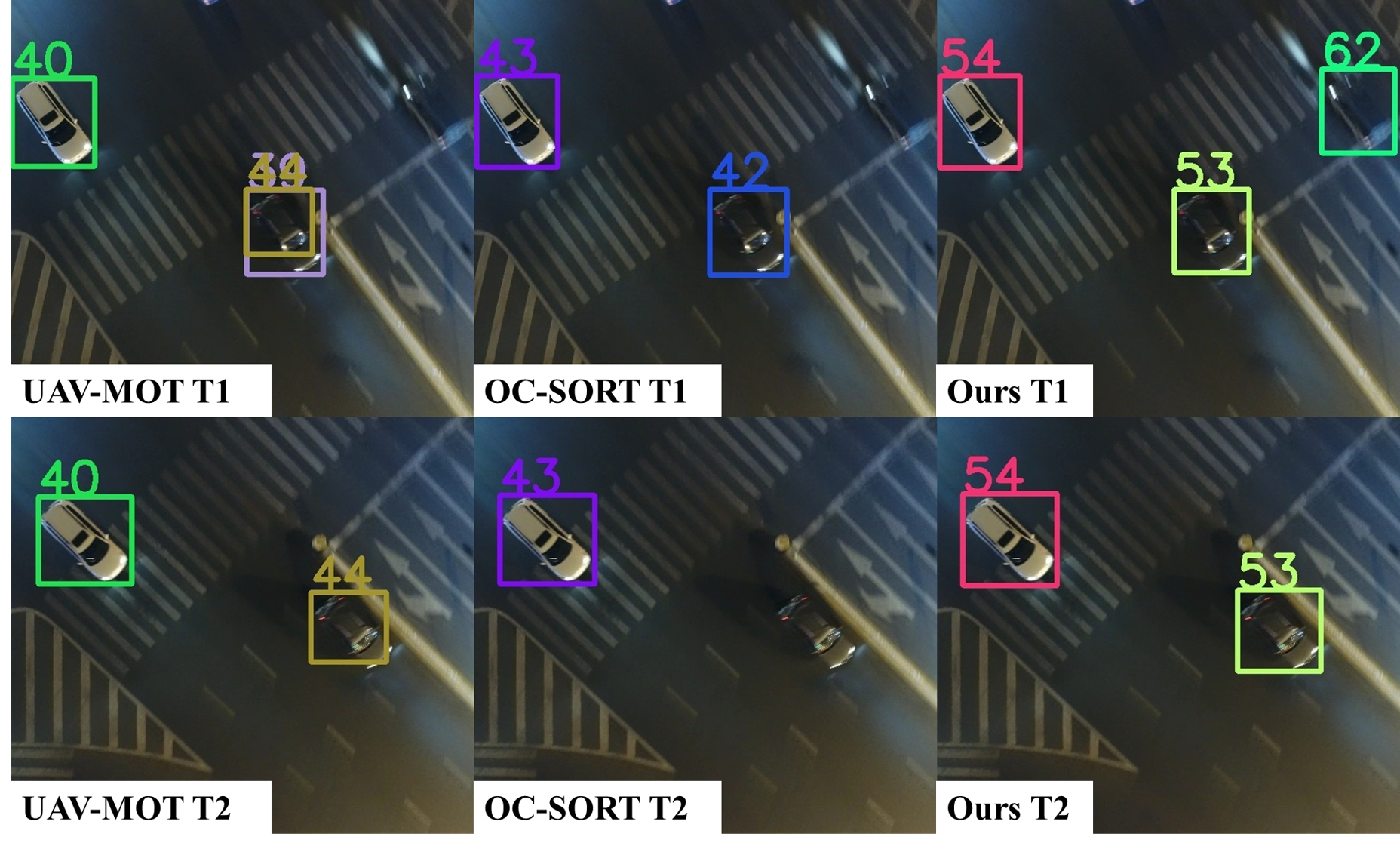}
 \label{fig:comparisons:uavdt}}
\caption{Visual comparisons of our method and state-of-the-arts on Visdrone and UAVDT datasets. The same number denotes the tracked same object in different frames. 
% (a) Our FOLT successfully tracks the small, blurry, and fast-moving electronic bicycle rider with ID 206, while UAVMOT and OCSORT missed it in both T1 and T2 frames. (b) Our FOLT successfully tracks the two blur and fast-moving cars with ID 62 and 53, while UAVMOT missed the 62 car at T1 and OCSORT missed the 53 car at T2. 
(a) Visual comparisons on the Visdrone dataset. (b) Visual comparisons on the UAVDT dataset.
}

\label{fig:comparisons}
\end{figure*}

\noindent\textbf{Combination.}
We also evaluate the effectiveness of our FGFA and FGMP which are used in combination.
Table~\ref{tab:ablation} shows that both FGFA and FGMP can effectively improve the tracking performance (39.6 to 40.9 and 39.6 to 40.9 in MOTA, 50.4 to 52.3 and 50.4 to 55.3 in IDF1).
With the FGFA and FGMP commonly used, the best tracking performance is obtained (42.1 in MOTA and 56.9 in IDF1), and the tracking speed is faster than the baseline model (27.0 to 29.4 in FPS).
% Therefore, our proposed FGFA and FGMP can not only improve tracking accuracy but also increase inference speed.

\subsection{Comparison with State-of-the-arts}
We conducted comparative experiments on the Visdrone test-dev set and the UAVDT test set.
Table~\ref{tab:sota_comparisons} shows the experimental results on these two datasets.
As shown in Table~\ref{tab:sota_comparisons}, the FOLT proposed in this paper exceeds the current state-of-the-art (SOTA) methods in several important metrics such as MOTA, IDF1, IDs, and MOTP.
For example, the MOTA of FOLT is 2.5\% higher than the previous highest method OCSORT (42.1 compared with 39.6)
, the IDF1 is 5.9\% higher than the previous highest method UAVMOT (56.9 compared with 51.0), while the inference speed is also faster than the previous fastest method ByteTrack (29.4 compared with 27.0).
% From column 8 to column 10 of Visdrone's results we can see that the big improvement of MOTA mainly comes from the reduction of false negatives (FN), since the reduction of FN compared with OCSORT is 15883, while the reduction of IDs 186 and FP increased 9474.
% This is because FOLT successfully detects and associates objects in fast motion and small objects, which is missed in the previous tracker.
FOLT also achieves the best MOTA and IDF1 in the UAVDT dataset, with MOTA 1.0\% higher than the previous best method OCSORT (48.5 compared with 47.5), and IDF1 1.0\% higher than the previous best method UAVMOT, (68.3 compared with 67.3)
showing the strongest ability in tracking objects from a UAV view.
In addition, our method that only uses location information in object matching also performs better than those that use both location and appearance information such as UAVMOT and SiamMOT, validating our belief that appearance feature is not reliable in the small object with a blurry appearance.

% To demonstrate the effectiveness of our proposed FOLT more intuitively, 
We also visualize the comparison results of the latest best methods (UAVMOT and OCSORT) and our FOLT in the Visdrone dataset and the UAVDT dataset.
As Fig~\ref{fig:comparisons:visdrone} shows, our FOLT successfully tracks the small and fast-moving electronic bicycle rider (with object ID 206) in both $T1$ and $T2$ frames, while the UAV-MOT and the OC-SORT tracker all failed to track it in both frames. 
Therefore, our FOLT is better than UAVMOT and OCSORT in tracking objects with small sizes and fast motion.
As Fig~\ref{fig:comparisons:uavdt} shows, the camera and the object are both in motion, introducing large and irregular motion and causing the object feature to blur.
In this difficult situation, our FOLT successfully tracks three cars in frame $T1$ with object ID 54, 53, and 62 respectively, while the UAVMOT has duplicate tracking on the middle car and the OCSORT missed the right-top car that is severely blurred due to large and irregular motion.
Our FOLT also successfully tracks two cars in frame $T2$ with object size 54 and 53, while the OC-SORT missed the fast-moving cars with ID 42.
In summary, our FOLT performs better than the state-of-the-arts in objects with small sizes, large and irregular motion, and blurred appearance.

% \vspace{0.01cm}

\section{Conclusion}
\label{chap:conclusion}
Multiple objects tracking in UAV view face the difficulty of large and irregular motion of both ground objects and UAV platforms. The small object size and blurred object appearance also hinder the tracking process.
In this paper, we propose FOLT to address these challenging problems.
The FOLT uses a modern light-weight detector and optical flow estimator to extract object detection features and motion information with high efficiency.
Given these extracted features, FOLT introduces the flow-guided feature augmentation (FGFA) to augment detection features based on previous features and optical flow, which improves the detection of small objects and blurred objects.
We then propose flow-guided motion prediction (FGMP) which utilizes optical flow and a convolution layer to track objects with large and irregular motion more precisely.
Experiments show that both FGFA and FGMP can improve the accuracy of multiple object tracking in UAV view, and the combination of the two methods further improves the accuracy and achieves the best results.
Comparison with state-of-the-arts shows that FOLT achieves the best results on the two public UAV-MOT datasets in both tracking accuracy and inference speed.
\section*{Acknowledgement}
This work was supported in part by Natural Science Foundation of
China under contract 62171139, and in part by Zhongshan science
and technology development project under contract 2020AG016.
\vfill\pagebreak

\bibliographystyle{ACM-Reference-Format}
\bibliography{sample-base}


\begin{thebibliography}{52}

%%% the Web address.

\ifx \showCODEN    \undefined \def \showCODEN     #1{\unskip}     \fi
\ifx \showDOI      \undefined \def \showDOI       #1{#1}\fi
\ifx \showISBNx    \undefined \def \showISBNx     #1{\unskip}     \fi
\ifx \showISBNxiii \undefined \def \showISBNxiii  #1{\unskip}     \fi
\ifx \showISSN     \undefined \def \showISSN      #1{\unskip}     \fi
\ifx \showLCCN     \undefined \def \showLCCN      #1{\unskip}     \fi
\ifx \shownote     \undefined \def \shownote      #1{#1}          \fi
\ifx \showarticletitle \undefined \def \showarticletitle #1{#1}   \fi
\ifx \showURL      \undefined \def \showURL       {\relax}        \fi
% invisible to TeX
\providecommand\bibfield[2]{#2}
\providecommand\bibinfo[2]{#2}
\providecommand\natexlab[1]{#1}
\providecommand\showeprint[2][]{arXiv:#2}

\bibitem[Bae and Yoon(2017)]%
        {bae2017confidence}
\bibfield{author}{\bibinfo{person}{Seung-Hwan Bae} {and}
  \bibinfo{person}{Kuk-Jin Yoon}.} \bibinfo{year}{2017}\natexlab{}.
\newblock \showarticletitle{Confidence-based data association and
  discriminative deep appearance learning for robust online multi-object
  tracking}.
\newblock \bibinfo{journal}{\emph{IEEE transactions on pattern analysis and
  machine intelligence}} \bibinfo{volume}{40}, \bibinfo{number}{3}
  (\bibinfo{year}{2017}), \bibinfo{pages}{595--610}.
\newblock


\bibitem[Bewley et~al\mbox{.}(2016)]%
        {bewley2016simple}
\bibfield{author}{\bibinfo{person}{Alex Bewley}, \bibinfo{person}{Zongyuan Ge},
  \bibinfo{person}{Lionel Ott}, \bibinfo{person}{Fabio Ramos}, {and}
  \bibinfo{person}{Ben Upcroft}.} \bibinfo{year}{2016}\natexlab{}.
\newblock \showarticletitle{Simple online and realtime tracking}. In
  \bibinfo{booktitle}{\emph{ICIP}}. IEEE, \bibinfo{pages}{3464--3468}.
\newblock


\bibitem[Bochinski et~al\mbox{.}(2017)]%
        {bochinski2017high}
\bibfield{author}{\bibinfo{person}{Erik Bochinski}, \bibinfo{person}{Volker
  Eiselein}, {and} \bibinfo{person}{Thomas Sikora}.}
  \bibinfo{year}{2017}\natexlab{}.
\newblock \showarticletitle{High-speed tracking-by-detection without using
  image information}. In \bibinfo{booktitle}{\emph{AVSS2017}}. IEEE,
  \bibinfo{pages}{1--6}.
\newblock


\bibitem[Brox et~al\mbox{.}(2006)]%
        {brox2006high}
\bibfield{author}{\bibinfo{person}{Thomas Brox}, \bibinfo{person}{Bodo
  Rosenhahn}, \bibinfo{person}{Daniel Cremers}, {and}
  \bibinfo{person}{Hans-Peter Seidel}.} \bibinfo{year}{2006}\natexlab{}.
\newblock \showarticletitle{High accuracy optical flow serves 3-D pose
  tracking: exploiting contour and flow based constraints}. In
  \bibinfo{booktitle}{\emph{ECCV}}. Springer, \bibinfo{pages}{98--111}.
\newblock


\bibitem[Cao et~al\mbox{.}(2022)]%
        {cao2022observation}
\bibfield{author}{\bibinfo{person}{Jinkun Cao}, \bibinfo{person}{Xinshuo Weng},
  \bibinfo{person}{Rawal Khirodkar}, \bibinfo{person}{Jiangmiao Pang}, {and}
  \bibinfo{person}{Kris Kitani}.} \bibinfo{year}{2022}\natexlab{}.
\newblock \showarticletitle{Observation-Centric SORT: Rethinking SORT for
  Robust Multi-Object Tracking}.
\newblock \bibinfo{journal}{\emph{arXiv preprint arXiv:2203.14360}}
  (\bibinfo{year}{2022}).
\newblock


\bibitem[Chandra et~al\mbox{.}(2015)]%
        {chandra2015eye}
\bibfield{author}{\bibinfo{person}{Sushil Chandra}, \bibinfo{person}{Greeshma
  Sharma}, \bibinfo{person}{Saloni Malhotra}, \bibinfo{person}{Devendra Jha},
  {and} \bibinfo{person}{Alok~Prakash Mittal}.}
  \bibinfo{year}{2015}\natexlab{}.
\newblock \showarticletitle{Eye tracking based human computer interaction:
  Applications and their uses}. In \bibinfo{booktitle}{\emph{2015 International
  Conference on Man and Machine Interfacing (MAMI)}}. IEEE,
  \bibinfo{pages}{1--5}.
\newblock


\bibitem[Chen et~al\mbox{.}(2019)]%
        {2019Deep}
\bibfield{author}{\bibinfo{person}{S. Chen}, \bibinfo{person}{Y. Xu},
  \bibinfo{person}{X. Zhou}, {and} \bibinfo{person}{F. Li}.}
  \bibinfo{year}{2019}\natexlab{}.
\newblock \showarticletitle{Deep Learning for Multiple Object Tracking: A
  Survey}.
\newblock \bibinfo{journal}{\emph{IET Computer Vision}} \bibinfo{volume}{13},
  \bibinfo{number}{4} (\bibinfo{year}{2019}), \bibinfo{pages}{355--368}.
\newblock


\bibitem[Dendorfer et~al\mbox{.}(2020)]%
        {dendorfer2020mot20}
\bibfield{author}{\bibinfo{person}{Patrick Dendorfer}, \bibinfo{person}{Hamid
  Rezatofighi}, \bibinfo{person}{Anton Milan}, \bibinfo{person}{Javen Shi},
  \bibinfo{person}{Daniel Cremers}, \bibinfo{person}{Ian Reid},
  \bibinfo{person}{Stefan Roth}, \bibinfo{person}{Konrad Schindler}, {and}
  \bibinfo{person}{Laura Leal-Taix{\'e}}.} \bibinfo{year}{2020}\natexlab{}.
\newblock \showarticletitle{Mot20: A benchmark for multi object tracking in
  crowded scenes}.
\newblock \bibinfo{journal}{\emph{arXiv preprint arXiv:2003.09003}}
  (\bibinfo{year}{2020}).
\newblock


\bibitem[Dosovitskiy et~al\mbox{.}(2015)]%
        {dosovitskiy2015flownet}
\bibfield{author}{\bibinfo{person}{Alexey Dosovitskiy},
  \bibinfo{person}{Philipp Fischer}, \bibinfo{person}{Eddy Ilg},
  \bibinfo{person}{Philip Hausser}, \bibinfo{person}{Caner Hazirbas},
  \bibinfo{person}{Vladimir Golkov}, \bibinfo{person}{Patrick Van Der~Smagt},
  \bibinfo{person}{Daniel Cremers}, {and} \bibinfo{person}{Thomas Brox}.}
  \bibinfo{year}{2015}\natexlab{}.
\newblock \showarticletitle{Flownet: Learning optical flow with convolutional
  networks}. In \bibinfo{booktitle}{\emph{Proceedings of the IEEE international
  conference on computer vision}}. \bibinfo{pages}{2758--2766}.
\newblock


\bibitem[Du et~al\mbox{.}(2018)]%
        {du2018unmanned}
\bibfield{author}{\bibinfo{person}{Dawei Du}, \bibinfo{person}{Yuankai Qi},
  \bibinfo{person}{Hongyang Yu}, \bibinfo{person}{Yifan Yang},
  \bibinfo{person}{Kaiwen Duan}, \bibinfo{person}{Guorong Li},
  \bibinfo{person}{Weigang Zhang}, \bibinfo{person}{Qingming Huang}, {and}
  \bibinfo{person}{Qi Tian}.} \bibinfo{year}{2018}\natexlab{}.
\newblock \showarticletitle{The unmanned aerial vehicle benchmark: Object
  detection and tracking}. In \bibinfo{booktitle}{\emph{ECCV}}.
  \bibinfo{pages}{370--386}.
\newblock


\bibitem[Fagot-Bouquet et~al\mbox{.}(2016)]%
        {fagot2016improving}
\bibfield{author}{\bibinfo{person}{Lo{\"\i}c Fagot-Bouquet},
  \bibinfo{person}{Romaric Audigier}, \bibinfo{person}{Yoann Dhome}, {and}
  \bibinfo{person}{Fr{\'e}d{\'e}ric Lerasle}.} \bibinfo{year}{2016}\natexlab{}.
\newblock \showarticletitle{Improving multi-frame data association with sparse
  representations for robust near-online multi-object tracking}. In
  \bibinfo{booktitle}{\emph{European Conference on Computer Vision}}. Springer,
  \bibinfo{pages}{774--790}.
\newblock


\bibitem[Forsyth et~al\mbox{.}(2006)]%
        {forsyth2006computational}
\bibfield{author}{\bibinfo{person}{David~A Forsyth}, \bibinfo{person}{Okan
  Arikan}, \bibinfo{person}{Leslie Ikemoto}, \bibinfo{person}{James O'Brien},
  \bibinfo{person}{Deva Ramanan}, {et~al\mbox{.}}}
  \bibinfo{year}{2006}\natexlab{}.
\newblock \showarticletitle{Computational studies of human motion: Part 1,
  tracking and motion synthesis}.
\newblock \bibinfo{journal}{\emph{Foundations and Trends{\textregistered} in
  Computer Graphics and Vision}} \bibinfo{volume}{1}, \bibinfo{number}{2--3}
  (\bibinfo{year}{2006}), \bibinfo{pages}{77--254}.
\newblock


\bibitem[Ge et~al\mbox{.}(2021)]%
        {ge2021yolox}
\bibfield{author}{\bibinfo{person}{Zheng Ge}, \bibinfo{person}{Songtao Liu},
  \bibinfo{person}{Feng Wang}, \bibinfo{person}{Zeming Li}, {and}
  \bibinfo{person}{Jian Sun}.} \bibinfo{year}{2021}\natexlab{}.
\newblock \showarticletitle{Yolox: Exceeding yolo series in 2021}.
\newblock \bibinfo{journal}{\emph{arXiv preprint arXiv:2107.08430}}
  (\bibinfo{year}{2021}).
\newblock


\bibitem[Geiger et~al\mbox{.}(2013)]%
        {geiger2013vision}
\bibfield{author}{\bibinfo{person}{Andreas Geiger}, \bibinfo{person}{Philip
  Lenz}, \bibinfo{person}{Christoph Stiller}, {and} \bibinfo{person}{Raquel
  Urtasun}.} \bibinfo{year}{2013}\natexlab{}.
\newblock \showarticletitle{Vision meets robotics: The kitti dataset}.
\newblock \bibinfo{journal}{\emph{The International Journal of Robotics
  Research}} \bibinfo{volume}{32}, \bibinfo{number}{11} (\bibinfo{year}{2013}),
  \bibinfo{pages}{1231--1237}.
\newblock


\bibitem[Geiger et~al\mbox{.}(2012)]%
        {geiger2012we}
\bibfield{author}{\bibinfo{person}{Andreas Geiger}, \bibinfo{person}{Philip
  Lenz}, {and} \bibinfo{person}{Raquel Urtasun}.}
  \bibinfo{year}{2012}\natexlab{}.
\newblock \showarticletitle{Are we ready for autonomous driving? the kitti
  vision benchmark suite}. In \bibinfo{booktitle}{\emph{2012 IEEE conference on
  computer vision and pattern recognition}}. IEEE, \bibinfo{pages}{3354--3361}.
\newblock


\bibitem[He et~al\mbox{.}(2016)]%
        {he2016deep}
\bibfield{author}{\bibinfo{person}{Kaiming He}, \bibinfo{person}{Xiangyu
  Zhang}, \bibinfo{person}{Shaoqing Ren}, {and} \bibinfo{person}{Jian Sun}.}
  \bibinfo{year}{2016}\natexlab{}.
\newblock \showarticletitle{Deep residual learning for image recognition}. In
  \bibinfo{booktitle}{\emph{Proceedings of the IEEE conference on computer
  vision and pattern recognition}}. \bibinfo{pages}{770--778}.
\newblock


\bibitem[Horn and Schunck(1981)]%
        {horn1981determining}
\bibfield{author}{\bibinfo{person}{Berthold~KP Horn} {and}
  \bibinfo{person}{Brian~G Schunck}.} \bibinfo{year}{1981}\natexlab{}.
\newblock \showarticletitle{Determining optical flow}.
\newblock \bibinfo{journal}{\emph{Artificial intelligence}}
  \bibinfo{volume}{17}, \bibinfo{number}{1-3} (\bibinfo{year}{1981}),
  \bibinfo{pages}{185--203}.
\newblock


\bibitem[Hur and Roth(2019)]%
        {hur2019iterative}
\bibfield{author}{\bibinfo{person}{Junhwa Hur} {and} \bibinfo{person}{Stefan
  Roth}.} \bibinfo{year}{2019}\natexlab{}.
\newblock \showarticletitle{Iterative residual refinement for joint optical
  flow and occlusion estimation}. In \bibinfo{booktitle}{\emph{Proceedings of
  the IEEE/CVF Conference on Computer Vision and Pattern Recognition}}.
  \bibinfo{pages}{5754--5763}.
\newblock


\bibitem[Janai et~al\mbox{.}(2020)]%
        {janai2020computer}
\bibfield{author}{\bibinfo{person}{Joel Janai}, \bibinfo{person}{Fatma
  G{\"u}ney}, \bibinfo{person}{Aseem Behl}, \bibinfo{person}{Andreas Geiger},
  {et~al\mbox{.}}} \bibinfo{year}{2020}\natexlab{}.
\newblock \showarticletitle{Computer vision for autonomous vehicles: Problems,
  datasets and state of the art}.
\newblock \bibinfo{journal}{\emph{Foundations and Trends{\textregistered} in
  Computer Graphics and Vision}} \bibinfo{volume}{12}, \bibinfo{number}{1--3}
  (\bibinfo{year}{2020}), \bibinfo{pages}{1--308}.
\newblock


\bibitem[Kalake et~al\mbox{.}(2021)]%
        {2021Analysis}
\bibfield{author}{\bibinfo{person}{L. Kalake}, \bibinfo{person}{W. Wan}, {and}
  \bibinfo{person}{L. Hou}.} \bibinfo{year}{2021}\natexlab{}.
\newblock \showarticletitle{Analysis Based on Recent Deep Learning Approaches
  Applied in Real-Time Multi-Object Tracking: A Review}.
\newblock \bibinfo{journal}{\emph{IEEE Access}} \bibinfo{volume}{PP},
  \bibinfo{number}{99} (\bibinfo{year}{2021}), \bibinfo{pages}{1--1}.
\newblock


\bibitem[Kalman(1960)]%
        {kalman1960new}
\bibfield{author}{\bibinfo{person}{Rudolph~Emil Kalman}.}
  \bibinfo{year}{1960}\natexlab{}.
\newblock \showarticletitle{A new approach to linear filtering and prediction
  problems}.
\newblock  (\bibinfo{year}{1960}).
\newblock


\bibitem[Kong et~al\mbox{.}(2021)]%
        {kong2021fastflownet}
\bibfield{author}{\bibinfo{person}{Lingtong Kong}, \bibinfo{person}{Chunhua
  Shen}, {and} \bibinfo{person}{Jie Yang}.} \bibinfo{year}{2021}\natexlab{}.
\newblock \showarticletitle{Fastflownet: A lightweight network for fast optical
  flow estimation}. In \bibinfo{booktitle}{\emph{ICRA}}. IEEE,
  \bibinfo{pages}{10310--10316}.
\newblock


\bibitem[Kratz and Nishino(2010)]%
        {kratz2010tracking}
\bibfield{author}{\bibinfo{person}{Louis Kratz} {and} \bibinfo{person}{Ko
  Nishino}.} \bibinfo{year}{2010}\natexlab{}.
\newblock \showarticletitle{Tracking with local spatio-temporal motion patterns
  in extremely crowded scenes}. In \bibinfo{booktitle}{\emph{2010 IEEE computer
  society conference on computer vision and pattern recognition}}. IEEE,
  \bibinfo{pages}{693--700}.
\newblock


\bibitem[Kuo et~al\mbox{.}(2010)]%
        {kuo2010multi}
\bibfield{author}{\bibinfo{person}{Cheng-Hao Kuo}, \bibinfo{person}{Chang
  Huang}, {and} \bibinfo{person}{Ramakant Nevatia}.}
  \bibinfo{year}{2010}\natexlab{}.
\newblock \showarticletitle{Multi-target tracking by on-line learned
  discriminative appearance models}. In \bibinfo{booktitle}{\emph{2010 IEEE
  Computer Society Conference on Computer Vision and Pattern Recognition}}.
  IEEE, \bibinfo{pages}{685--692}.
\newblock


\bibitem[Kurimo et~al\mbox{.}(2009)]%
        {kurimo2009effect}
\bibfield{author}{\bibinfo{person}{Eero Kurimo}, \bibinfo{person}{Leena
  Lepist{\"o}}, \bibinfo{person}{Jarno Nikkanen}, \bibinfo{person}{Juuso
  Gr{\'e}n}, \bibinfo{person}{Iivari Kunttu}, {and} \bibinfo{person}{Jorma
  Laaksonen}.} \bibinfo{year}{2009}\natexlab{}.
\newblock \showarticletitle{The effect of motion blur and signal noise on image
  quality in low light imaging}. In \bibinfo{booktitle}{\emph{Image Analysis:
  16th Scandinavian Conference, SCIA 2009, Oslo, Norway, June 15-18, 2009.
  Proceedings 16}}. Springer, \bibinfo{pages}{81--90}.
\newblock


\bibitem[Leal-Taix{\'e} et~al\mbox{.}(2015)]%
        {leal2015motchallenge}
\bibfield{author}{\bibinfo{person}{Laura Leal-Taix{\'e}},
  \bibinfo{person}{Anton Milan}, \bibinfo{person}{Ian Reid},
  \bibinfo{person}{Stefan Roth}, {and} \bibinfo{person}{Konrad Schindler}.}
  \bibinfo{year}{2015}\natexlab{}.
\newblock \showarticletitle{Motchallenge 2015: Towards a benchmark for
  multi-target tracking}.
\newblock \bibinfo{journal}{\emph{arXiv preprint arXiv:1504.01942}}
  (\bibinfo{year}{2015}).
\newblock


\bibitem[Li et~al\mbox{.}(2013)]%
        {li2013survey}
\bibfield{author}{\bibinfo{person}{Xi Li}, \bibinfo{person}{Weiming Hu},
  \bibinfo{person}{Chunhua Shen}, \bibinfo{person}{Zhongfei Zhang},
  \bibinfo{person}{Anthony Dick}, {and} \bibinfo{person}{Anton Van~Den
  Hengel}.} \bibinfo{year}{2013}\natexlab{}.
\newblock \showarticletitle{A survey of appearance models in visual object
  tracking}.
\newblock \bibinfo{journal}{\emph{ACM transactions on Intelligent Systems and
  Technology (TIST)}} \bibinfo{volume}{4}, \bibinfo{number}{4}
  (\bibinfo{year}{2013}), \bibinfo{pages}{1--48}.
\newblock


\bibitem[Liu et~al\mbox{.}(2022)]%
        {liu2022multi}
\bibfield{author}{\bibinfo{person}{Shuai Liu}, \bibinfo{person}{Xin Li},
  \bibinfo{person}{Huchuan Lu}, {and} \bibinfo{person}{You He}.}
  \bibinfo{year}{2022}\natexlab{}.
\newblock \showarticletitle{Multi-Object Tracking Meets Moving UAV}. In
  \bibinfo{booktitle}{\emph{Proceedings of the IEEE/CVF Conference on Computer
  Vision and Pattern Recognition}}. \bibinfo{pages}{8876--8885}.
\newblock


\bibitem[Luo et~al\mbox{.}(2021)]%
        {luo2021multiple}
\bibfield{author}{\bibinfo{person}{Wenhan Luo}, \bibinfo{person}{Junliang
  Xing}, \bibinfo{person}{Anton Milan}, \bibinfo{person}{Xiaoqin Zhang},
  \bibinfo{person}{Wei Liu}, {and} \bibinfo{person}{Tae-Kyun Kim}.}
  \bibinfo{year}{2021}\natexlab{}.
\newblock \showarticletitle{Multiple object tracking: A literature review}.
\newblock \bibinfo{journal}{\emph{Artificial Intelligence}}
  \bibinfo{volume}{293} (\bibinfo{year}{2021}), \bibinfo{pages}{103448}.
\newblock


\bibitem[Meinhardt et~al\mbox{.}(2021)]%
        {2021TrackFormer}
\bibfield{author}{\bibinfo{person}{T. Meinhardt}, \bibinfo{person}{A.
  Kirillov}, \bibinfo{person}{L. Leal-Taixe}, {and} \bibinfo{person}{C.
  Feichtenhofer}.} \bibinfo{year}{2021}\natexlab{}.
\newblock \showarticletitle{TrackFormer: Multi-Object Tracking with
  Transformers}.
\newblock  (\bibinfo{year}{2021}).
\newblock


\bibitem[Milan et~al\mbox{.}(2017)]%
        {milan2017online}
\bibfield{author}{\bibinfo{person}{Anton Milan}, \bibinfo{person}{S~Hamid
  Rezatofighi}, \bibinfo{person}{Anthony Dick}, \bibinfo{person}{Ian Reid},
  {and} \bibinfo{person}{Konrad Schindler}.} \bibinfo{year}{2017}\natexlab{}.
\newblock \showarticletitle{Online multi-target tracking using recurrent neural
  networks}. In \bibinfo{booktitle}{\emph{AAAI}}, Vol.~\bibinfo{volume}{31}.
\newblock


\bibitem[Peng et~al\mbox{.}(2020a)]%
        {peng2020dense}
\bibfield{author}{\bibinfo{person}{Jinlong Peng}, \bibinfo{person}{Yueyang Gu},
  \bibinfo{person}{Yabiao Wang}, \bibinfo{person}{Chengjie Wang},
  \bibinfo{person}{Jilin Li}, {and} \bibinfo{person}{Feiyue Huang}.}
  \bibinfo{year}{2020}\natexlab{a}.
\newblock \showarticletitle{Dense scene multiple object tracking with box-plane
  matching}. In \bibinfo{booktitle}{\emph{ACM International Conference on
  Multimedia}}. \bibinfo{pages}{4615--4619}.
\newblock


\bibitem[Peng et~al\mbox{.}(2018)]%
        {peng2018tracklet}
\bibfield{author}{\bibinfo{person}{Jinlong Peng}, \bibinfo{person}{Fan Qiu},
  \bibinfo{person}{John See}, \bibinfo{person}{Qi Guo},
  \bibinfo{person}{Shaoshuai Huang}, \bibinfo{person}{Ling-Yu Duan}, {and}
  \bibinfo{person}{Weiyao Lin}.} \bibinfo{year}{2018}\natexlab{}.
\newblock \showarticletitle{Tracklet siamese network with constrained
  clustering for multiple object tracking}. In
  \bibinfo{booktitle}{\emph{VCIP}}. IEEE, \bibinfo{pages}{1--4}.
\newblock


\bibitem[Peng et~al\mbox{.}(2020c)]%
        {peng2020chained}
\bibfield{author}{\bibinfo{person}{Jinlong Peng}, \bibinfo{person}{Changan
  Wang}, \bibinfo{person}{Fangbin Wan}, \bibinfo{person}{Yang Wu},
  \bibinfo{person}{Yabiao Wang}, \bibinfo{person}{Ying Tai},
  \bibinfo{person}{Chengjie Wang}, \bibinfo{person}{Jilin Li},
  \bibinfo{person}{Feiyue Huang}, {and} \bibinfo{person}{Yanwei Fu}.}
  \bibinfo{year}{2020}\natexlab{c}.
\newblock \showarticletitle{Chained-tracker: Chaining paired attentive
  regression results for end-to-end joint multiple-object detection and
  tracking}. In \bibinfo{booktitle}{\emph{ECCV}}. Springer,
  \bibinfo{pages}{145--161}.
\newblock


\bibitem[Peng et~al\mbox{.}(2020b)]%
        {peng2020tpm}
\bibfield{author}{\bibinfo{person}{Jinlong Peng}, \bibinfo{person}{Tao Wang},
  \bibinfo{person}{Weiyao Lin}, \bibinfo{person}{Jian Wang},
  \bibinfo{person}{John See}, \bibinfo{person}{Shilei Wen}, {and}
  \bibinfo{person}{Erui Ding}.} \bibinfo{year}{2020}\natexlab{b}.
\newblock \showarticletitle{TPM: Multiple object tracking with tracklet-plane
  matching}.
\newblock \bibinfo{journal}{\emph{Pattern Recognition}}  \bibinfo{volume}{107}
  (\bibinfo{year}{2020}), \bibinfo{pages}{107480}.
\newblock


\bibitem[Pirsiavash et~al\mbox{.}(2011)]%
        {pirsiavash2011globally}
\bibfield{author}{\bibinfo{person}{Hamed Pirsiavash}, \bibinfo{person}{Deva
  Ramanan}, {and} \bibinfo{person}{Charless~C Fowlkes}.}
  \bibinfo{year}{2011}\natexlab{}.
\newblock \showarticletitle{Globally-optimal greedy algorithms for tracking a
  variable number of objects}. In \bibinfo{booktitle}{\emph{CVPR 2011}}. IEEE,
  \bibinfo{pages}{1201--1208}.
\newblock


\bibitem[Shuai et~al\mbox{.}(2021)]%
        {shuai2021siammot}
\bibfield{author}{\bibinfo{person}{Bing Shuai}, \bibinfo{person}{Andrew
  Berneshawi}, \bibinfo{person}{Xinyu Li}, \bibinfo{person}{Davide Modolo},
  {and} \bibinfo{person}{Joseph Tighe}.} \bibinfo{year}{2021}\natexlab{}.
\newblock \showarticletitle{SiamMOT: Siamese Multi-Object Tracking}. In
  \bibinfo{booktitle}{\emph{Proceedings of the IEEE/CVF Conference on Computer
  Vision and Pattern Recognition}}. \bibinfo{pages}{12372--12382}.
\newblock


\bibitem[Simonyan and Zisserman(2014)]%
        {simonyan2014two}
\bibfield{author}{\bibinfo{person}{Karen Simonyan} {and}
  \bibinfo{person}{Andrew Zisserman}.} \bibinfo{year}{2014}\natexlab{}.
\newblock \showarticletitle{Two-stream convolutional networks for action
  recognition in videos}.
\newblock \bibinfo{journal}{\emph{Advances in neural information processing
  systems}}  \bibinfo{volume}{27} (\bibinfo{year}{2014}).
\newblock


\bibitem[Sugimura et~al\mbox{.}(2009)]%
        {sugimura2009using}
\bibfield{author}{\bibinfo{person}{Daisuke Sugimura}, \bibinfo{person}{Kris~M
  Kitani}, \bibinfo{person}{Takahiro Okabe}, \bibinfo{person}{Yoichi Sato},
  {and} \bibinfo{person}{Akihiro Sugimoto}.} \bibinfo{year}{2009}\natexlab{}.
\newblock \showarticletitle{Using individuality to track individuals:
  Clustering individual trajectories in crowds using local appearance and
  frequency trait}. In \bibinfo{booktitle}{\emph{ICCV}}. IEEE,
  \bibinfo{pages}{1467--1474}.
\newblock


\bibitem[Sun et~al\mbox{.}(2014)]%
        {sun2014quantitative}
\bibfield{author}{\bibinfo{person}{Deqing Sun}, \bibinfo{person}{Stefan Roth},
  {and} \bibinfo{person}{Michael~J Black}.} \bibinfo{year}{2014}\natexlab{}.
\newblock \showarticletitle{A quantitative analysis of current practices in
  optical flow estimation and the principles behind them}.
\newblock \bibinfo{journal}{\emph{International Journal of Computer Vision}}
  \bibinfo{volume}{106}, \bibinfo{number}{2} (\bibinfo{year}{2014}),
  \bibinfo{pages}{115--137}.
\newblock


\bibitem[Sun et~al\mbox{.}(2018)]%
        {sun2018pwc}
\bibfield{author}{\bibinfo{person}{Deqing Sun}, \bibinfo{person}{Xiaodong
  Yang}, \bibinfo{person}{Ming-Yu Liu}, {and} \bibinfo{person}{Jan Kautz}.}
  \bibinfo{year}{2018}\natexlab{}.
\newblock \showarticletitle{Pwc-net: Cnns for optical flow using pyramid,
  warping, and cost volume}. In \bibinfo{booktitle}{\emph{CVPR}}.
  \bibinfo{pages}{8934--8943}.
\newblock


\bibitem[Takala and Pietikainen(2007)]%
        {takala2007multi}
\bibfield{author}{\bibinfo{person}{Valtteri Takala} {and}
  \bibinfo{person}{Matti Pietikainen}.} \bibinfo{year}{2007}\natexlab{}.
\newblock \showarticletitle{Multi-object tracking using color, texture and
  motion}. In \bibinfo{booktitle}{\emph{2007 IEEE Conference on Computer Vision
  and Pattern Recognition}}. IEEE, \bibinfo{pages}{1--7}.
\newblock


\bibitem[Wang et~al\mbox{.}(2020)]%
        {wang2020towards}
\bibfield{author}{\bibinfo{person}{Zhongdao Wang}, \bibinfo{person}{Liang
  Zheng}, \bibinfo{person}{Yixuan Liu}, \bibinfo{person}{Yali Li}, {and}
  \bibinfo{person}{Shengjin Wang}.} \bibinfo{year}{2020}\natexlab{}.
\newblock \showarticletitle{Towards real-time multi-object tracking}. In
  \bibinfo{booktitle}{\emph{European Conference on Computer Vision}}. Springer,
  \bibinfo{pages}{107--122}.
\newblock


\bibitem[Wojke et~al\mbox{.}(2017)]%
        {wojke2017simple}
\bibfield{author}{\bibinfo{person}{Nicolai Wojke}, \bibinfo{person}{Alex
  Bewley}, {and} \bibinfo{person}{Dietrich Paulus}.}
  \bibinfo{year}{2017}\natexlab{}.
\newblock \showarticletitle{Simple online and realtime tracking with a deep
  association metric}. In \bibinfo{booktitle}{\emph{ICIP}}. IEEE,
  \bibinfo{pages}{3645--3649}.
\newblock


\bibitem[Xu et~al\mbox{.}(2020)]%
        {xu2020train}
\bibfield{author}{\bibinfo{person}{Yihong Xu}, \bibinfo{person}{Aljosa Osep},
  \bibinfo{person}{Yutong Ban}, \bibinfo{person}{Radu Horaud},
  \bibinfo{person}{Laura Leal-Taix{\'e}}, {and} \bibinfo{person}{Xavier
  Alameda-Pineda}.} \bibinfo{year}{2020}\natexlab{}.
\newblock \showarticletitle{How to train your deep multi-object tracker}. In
  \bibinfo{booktitle}{\emph{CVPR}}. \bibinfo{pages}{6787--6796}.
\newblock


\bibitem[Yang and Nevatia(2012)]%
        {yang2012multi}
\bibfield{author}{\bibinfo{person}{Bo Yang} {and} \bibinfo{person}{Ram
  Nevatia}.} \bibinfo{year}{2012}\natexlab{}.
\newblock \showarticletitle{Multi-target tracking by online learning of
  non-linear motion patterns and robust appearance models}. In
  \bibinfo{booktitle}{\emph{CVPR}}. IEEE, \bibinfo{pages}{1918--1925}.
\newblock


\bibitem[Zhang et~al\mbox{.}(2022)]%
        {zhang2022bytetrack}
\bibfield{author}{\bibinfo{person}{Yifu Zhang}, \bibinfo{person}{Peize Sun},
  \bibinfo{person}{Yi Jiang}, \bibinfo{person}{Dongdong Yu},
  \bibinfo{person}{Fucheng Weng}, \bibinfo{person}{Zehuan Yuan},
  \bibinfo{person}{Ping Luo}, \bibinfo{person}{Wenyu Liu}, {and}
  \bibinfo{person}{Xinggang Wang}.} \bibinfo{year}{2022}\natexlab{}.
\newblock \showarticletitle{Bytetrack: Multi-object tracking by associating
  every detection box}. In \bibinfo{booktitle}{\emph{ECCV}}. Springer,
  \bibinfo{pages}{1--21}.
\newblock


\bibitem[Zhang et~al\mbox{.}(2021)]%
        {zhang2021fairmot}
\bibfield{author}{\bibinfo{person}{Yifu Zhang}, \bibinfo{person}{Chunyu Wang},
  \bibinfo{person}{Xinggang Wang}, \bibinfo{person}{Wenjun Zeng}, {and}
  \bibinfo{person}{Wenyu Liu}.} \bibinfo{year}{2021}\natexlab{}.
\newblock \showarticletitle{Fairmot: On the fairness of detection and
  re-identification in multiple object tracking}.
\newblock \bibinfo{journal}{\emph{International Journal of Computer Vision}}
  \bibinfo{volume}{129}, \bibinfo{number}{11} (\bibinfo{year}{2021}),
  \bibinfo{pages}{3069--3087}.
\newblock


\bibitem[Zhao et~al\mbox{.}(2012)]%
        {zhao2012tracking}
\bibfield{author}{\bibinfo{person}{Xuemei Zhao}, \bibinfo{person}{Dian Gong},
  {and} \bibinfo{person}{G{\'e}rard Medioni}.} \bibinfo{year}{2012}\natexlab{}.
\newblock \showarticletitle{Tracking using motion patterns for very crowded
  scenes}. In \bibinfo{booktitle}{\emph{ECCV}}. Springer,
  \bibinfo{pages}{315--328}.
\newblock


\bibitem[Zheng et~al\mbox{.}(2016)]%
        {zheng2016mars}
\bibfield{author}{\bibinfo{person}{Liang Zheng}, \bibinfo{person}{Zhi Bie},
  \bibinfo{person}{Yifan Sun}, \bibinfo{person}{Jingdong Wang},
  \bibinfo{person}{Chi Su}, \bibinfo{person}{Shengjin Wang}, {and}
  \bibinfo{person}{Qi Tian}.} \bibinfo{year}{2016}\natexlab{}.
\newblock \showarticletitle{Mars: A video benchmark for large-scale person
  re-identification}. In \bibinfo{booktitle}{\emph{European conference on
  computer vision}}. Springer, \bibinfo{pages}{868--884}.
\newblock


\bibitem[Zhou et~al\mbox{.}(2020)]%
        {zhou2020tracking}
\bibfield{author}{\bibinfo{person}{Xingyi Zhou}, \bibinfo{person}{Vladlen
  Koltun}, {and} \bibinfo{person}{Philipp Kr{\"a}henb{\"u}hl}.}
  \bibinfo{year}{2020}\natexlab{}.
\newblock \showarticletitle{Tracking objects as points}. In
  \bibinfo{booktitle}{\emph{European Conference on Computer Vision}}. Springer,
  \bibinfo{pages}{474--490}.
\newblock


\bibitem[Zhu et~al\mbox{.}(2020)]%
        {zhu2020detection}
\bibfield{author}{\bibinfo{person}{Pengfei Zhu}, \bibinfo{person}{Longyin Wen},
  \bibinfo{person}{Dawei Du}, \bibinfo{person}{Xiao Bian},
  \bibinfo{person}{Heng Fan}, \bibinfo{person}{Qinghua Hu}, {and}
  \bibinfo{person}{Haibin Ling}.} \bibinfo{year}{2020}\natexlab{}.
\newblock \showarticletitle{Detection and tracking meet drones challenge}.
\newblock \bibinfo{journal}{\emph{arXiv preprint arXiv:2001.06303}}
  (\bibinfo{year}{2020}).
\newblock


\end{thebibliography}
% {\small

% }

\end{document}